\documentclass[letterpaper]{article}
\usepackage{spconf,amsmath,graphicx,amsfonts,amssymb,amsthm}
\usepackage{algorithm,algpseudocode}
\usepackage{float}
\usepackage{tabularx,ragged2e,booktabs,array}
\usepackage{comment}

\def\one{{\mathbf 1}}

\title{A new geometric approach to latent topic modeling and discovery}
\name{Weicong Ding,  Mohammad H. Rohban,  Prakash Ishwar, Venkatesh Saligrama
}

\address{Department of Electrical and Computer Engineering, Boston
  University, Boston, MA, USA. }
\begin{document}
%\ninept
%
\maketitle
\begin{abstract}
A new geometrically-motivated algorithm for nonnegative matrix
factorization is developed and applied to the discovery of latent
``topics'' for text and image ``document'' corpora.  The algorithm is
based on robustly finding and clustering extreme-points of empirical
cross-document word-frequencies that correspond to novel ``words''
unique to each topic. In contrast to related approaches that are based
on solving non-convex optimization problems using suboptimal
approximations, locally-optimal methods, or heuristics, the new
algorithm is convex, has polynomial complexity, and has competitive
qualitative and quantitative performance compared to the current
state-of-the-art approaches on synthetic and real-world datasets.
\end{abstract}
\begin{keywords}
Topic modeling, nonnegative matrix factorization (NMF), extreme
points, subspace clustering.
\end{keywords}
%\vfill\pagebreak
%
\section{Introduction}
\label{sec:intro}

%Context, motivation, application.
Topic modeling is a statistical tool for the automatic discovery and
comprehension of latent thematic structure or {\it topics}, assumed to
pervade a corpus of documents.

Suppose that we have a corpus of $M$ documents composed of words from
a vocabulary of $W$ distinct words indexed by $w = 1,\ldots,W$. In the
classic ``bags of words'' modeling paradigm widely-used in
Probabilistic Latent Semantic Analysis \cite{PLSA99} and Latent
Dirichlet Allocation (LDA) \cite{LDA:ref,Blei2012Review:ref}, each
document is modeled as being generated by $N$ independent and
identically distributed (iid) drawings of words from an unknown
$W\times 1$ document word-distribution vector. Each document
word-distribution vector is itself modeled as an unknown {\it
  probabilistic mixture} of $K < \min(M,W)$ unknown $W \times 1$
latent topic word-distribution vectors that are {\it shared} among the
$M$ documents in the corpus. The goal of topic modeling then is to
estimate the latent topic word-distribution vectors and possibly the
topic mixing weights for each document from the empirical
word-frequency vectors of all documents.
Topic modeling has also been applied to various types of data other
than text, e.g., images, videos (with photometric and spatio-temporal
feature-vectors interpreted as the words), genetic sequences,
hyper-spectral images, voice, and music, for signal separation and
blind deconvolution.

If $\beta$ denotes the unknown $W \times K$ topic-matrix whose columns
are the $K$ latent topic word-distribution vectors and $\theta$
denotes the $K \times M$ weight-matrix whose $M$ columns are the
mixing weights over $K$ topics for the $M$ documents, then each column
of the $W \times M$ matrix $A = \beta\theta$ corresponds to a document
word-distribution vector. Let $X$ denote the observed $W \times M$
words-by-documents matrix whose $M$ columns are the {\it empirical}
word-frequency vectors of the $M$ documents when each document is
generated by $N$ iid drawings of words from the corresponding column
of the $A$ matrix. Then given only $X$ and $K$, the goal is to
estimate the topic matrix $\beta$ and possibly the weight-matrix
$\theta$. This can be formulated as a nonnegative matrix factorization
(NMF) problem
\cite{nmfLS:ref,Donhunique:ref,NMFbook:ref,recht2012factoring}
%
%\cite{convexNMF:ref}
%
where the typical solution strategy is to minimize a cost function of
the form
%
%\vspace{0ex}
% 
\begin{eqnarray}
%
%&& \nonumber \\ [-6ex]
%
\|X - \beta \theta \|^2 &+& \psi(\beta,\theta)
\label{eq:nmf}
\end{eqnarray}
%
%\vspace{0ex}
%
where the regularization term $\psi$ is introduced to enforce
desirable properties in the solution such as uniqueness of the
factorization, sparsity, etc. The joint optimization of (\ref{eq:nmf})
with respect to $(\beta,\theta)$ is, however, non-convex and
necessitates the use of suboptimal strategies such as alternating
minimization, greedy gradient descent, local search, approximations,
and heuristics. These are also typically sensitive to small sample
sizes (words per document) $N$ especially when $N \ll W$ because many
words may not be sampled and $X$ may be far from $A$ in Euclidean
distance. In LDA, the columns of $\beta$ and $\theta$ are modeled as
iid random drawings from Dirichlet {\it prior} distributions. The
resulting maximum aposteriori probability estimation of
$(\beta,\theta)$, however, turns out to be a fairly complex non-convex
problem. One then takes recourse to sub-optimal solutions based on
variational Bayes approximations of the posterior distribution and
other methods based on Gibbs sampling and expectation propagation.

In contrast to these approaches we adopt the non-negative matrix
factorization framework and propose a new geometrically motivated
algorithm that has competitive performance compared to the current
state-of-the art and is free of heuristics and approximations.

\section{A new geometric approach}
\label{sec:Geoprop}
%
% reasons to make separable Assumption ASM1

\begin{figure}[!htb]
\begin{minipage}[b]{1.0\linewidth}
 \centering
  \centerline{\includegraphics[width=8cm]{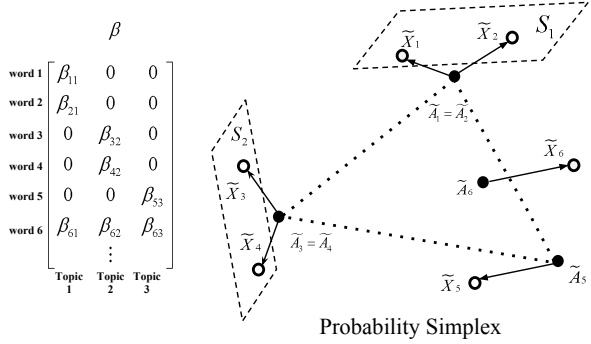}}
%  \vspace{2.0cm}
% \centerline{(a) Result 1}\medskip
\end{minipage}
\vglue -4ex
\caption{A separable topic matrix and the underlying geometric
  structure. Solid circles represent rows of $\widetilde{A}$, empty
  circles represent rows of $\widetilde{X}$.}
\label{fig:extreme}
\end{figure}

A key ingredient of the new approach is the so-called ``separability''
assumption introduced in \cite{Donhunique:ref} to ensure the
uniqueness of nonnegative matrix factorization. Applied to $\beta$
this means that each topic contains ``novel'' words which appear only
in that topic -- a property that has been found to hold in the
estimates of topic matrices produced by several algorithms
\cite{ARORA:ref}. More precisely,
%
%\begin{asm}{(Separability)}
A $W\times K$ topic matrix $\beta$ is separable if for each $k \in
[1,K]$, there exists a row of $\beta$ that has a single non-zero entry
which is in the $k$-th column.
%\end{asm}
%
\noindent Figure~\ref{fig:extreme} shows an example of a separable
topic matrix with three topics. Words 1 and 2 are unique (novel) to
topic 1, words 3, 4 to topic 2, and word 5 to topic 3.

Let $\mathcal{C}_k$ be the set of novel words of topic $k$ for $ k \in
[1,K]$ and let $\mathcal{C}_0$ be the remaining words in the
vocabulary. Let $A_w$ and $\theta_k$ denote the $w$-th and $k$-th {\it
  row}-vectors of $A$ and $\theta$ respectively. Observe that all the
row-vectors of $A$ that correspond to the novel words of the same
topic are just different scaled versions of the same $\theta$
row-vector: for each $w \in \mathcal{C}_k$, $A_w = \beta_{wk}
\theta_k$. Thus if $\widetilde{A}$, $\widetilde{\beta}$, and
$\widetilde{\theta}$ denote the {\it row-normalized} versions (i.e.,
unit row sums) of $A$, $\beta$, and $\theta$ respectively then
$\widetilde{A} = \widetilde{\beta}\widetilde{\theta}$ and for all $w
\in \mathcal{C}_k, \widetilde{A}_w = \widetilde{\theta}_k$ (e.g., in
Fig.~\ref{fig:extreme}, $\widetilde{A}_1 = \widetilde{A}_2 =
\widetilde{\theta}_1$ and $\widetilde{A}_3 = \widetilde{A}_4 =
\widetilde{\theta}_2$), and for all $w \in \mathcal{C}_0$,
$\widetilde{A}_w$ lives in the convex hull of $\widetilde{\theta}_k$'s
(in Fig.~\ref{fig:extreme}, $\widetilde{A}_6$ is in the convex hull of
$\widetilde{\theta}_1, \widetilde{\theta}_2, \widetilde{\theta}_3$).
%
%
%We normalize each row of $A, X, \theta$ as
%$\widetilde{A},\widetilde{X},\widetilde{\theta}$ respectively to be
%row stochastic, and let $\widetilde{\beta}=Diag(X\vec{1})\beta
%Diag(\theta\vec{1})^{-1})$ which is also row stochastic. Then
%$\widetilde{A} = \widetilde{\beta} \widetilde{\theta}$ while
%$\widetilde{X}$ is an estimation of $\widetilde{A}$. Each row-vector
%$\widetilde{X}_w$ or $\widetilde{A}_w$ is a point on probability
%simplex. Simple algebra shows that:
%
% properties derived from separable conditions.
%\begin{enumerate}
%
%For $w \in \mathcal{C}_k, k=1,\ldots,K$, $\widetilde{A}_w =
%\widetilde{\theta}_k$. For $w \in \mathcal{C}_0$, $\widetilde{A}_w, $
%lives in the convex hull of $\widetilde{\theta}_k$.
%
%The combination coefficients are the corresponding rows of $\widetilde{\beta}$.
%\end{enumerate}
%
\vglue -4ex
\vglue -2ex
%
%relation to previous work.
%
This geometric viewpoint reveals how to extract the topic matrix
$\beta$
%and $\theta$
from $A$: (1) Row-normalize $A$ to $\widetilde{A}$. (2) Find extreme
points of $\widetilde{A}$'s row-vectors. (3) Cluster the row-vectors of
$\widetilde{A}$ that correspond to the same extreme point into the
same group. There will be $K$ disjoint groups and each group will
correspond to the novel words of the same topic. (4) Express the
remaining row-vectors of $\widetilde{A}$ as convex combinations of the
extreme points. This gives us $\widetilde{\beta}$ 
%and $\widetilde{\theta}$. 
(5) Finally, renormalize $\widetilde{\beta}$ to obtain $\beta$.
%$\widetilde{A}$, $\widetilde{\beta}$ , and $\widetilde{\theta}$ to
%make them column-stochastic and give us $\beta$ and $\theta$.

The reality, however, is that we only have access to $X$, not $A$. The
above algorithm when applied to $X$ would work well if $X$ is close to
$A$ which would happen if $N$ is large. When $N$ is small, two
problems arise: (i) Points corresponding to novel words of the same
topic may become multiple extreme points and may be far from each
other (e.g., $\widetilde{X}_1, \widetilde{X}_2$ and $\widetilde{X}_3,
\widetilde{X}_4$ in Fig.~\ref{fig:extreme}). (ii) Points in the convex
hull may also become ``outlier'' extreme points (e.g.,
$\widetilde{X}_6$ in Fig.~\ref{fig:extreme}).

As a step towards overcoming these difficulties we observe that in
practice, the unique words of any topic only occur in a few
documents. This implies that the rows of $\theta$ are sparse and that
the row-vectors of $\widetilde{X}$ corresponding to the novel words of
the same topic are likely to form a {\it low-dimensional subspace}
(e.g., $S_1,S_2$ in Fig.~\ref{fig:extreme}) since their supports are
subsets of the supports of the same row-vector of $\theta$. If we make
the further assumption that for any pair of distinct topics there are
several documents in which their novel words do not {\it co-occur}
then the row subspaces of $\widetilde{X}$ corresponding to the novel
words any two distinct topics are likely to be significantly disjoint
(although they might share a common low-dimensional
subspace). Finally, the row-vectors of $\widetilde{X}$ corresponding
to non-novel words are unlikely to be close to the row subspaces of
$\widetilde{X}$ corresponding to the novel words any one topic (e.g.,
$\widetilde{X}_6$ in Fig.~\ref{fig:extreme}).
%All these properties hold even when we have few samples.
These observations and assumptions motivate the revised $5$-step
Algorithm~\ref{alg1} for extracting $\beta$
%and $\theta$
from $X$.
\begin{algorithm}[!htb]
\caption{Topic Discovery}
\label{alg1}
\begin{algorithmic}[1]
%with line numbers
%
% input
%
    \Require $W \times M$ word-document matrix $X$; \#~topics $K$.
%
% output
%
    \Ensure Estimate $\widehat{\beta}$ of $W\times K$ topic matrix
    $\beta$.
%
% steps
%
    \State Row-normalize $X$ to get $\widetilde{X}$.  Let $N_w :=
    \sum_{d=1}^{M}{X}_{wd}$.
    \State Apply Algorithm~\ref{randproj} to rows of $\widetilde{X}$
    to obtain a subset of rows $\mathcal{E}$ that correspond to {\it
      candidate} novel words. Let $\widehat{\mathcal{C}}_0$ be the
    remaining row indices.
%
% extracting extreme points
%
    \State Apply the sparse subspace clustering algorithm of
    \cite{SSC_candies:ref,SSC_vidal:ref}
% Algorithm \ref{SSC} 
    to $\mathcal{E}$ with parameters $\lambda_1,\gamma$ to obtain $K$
    clusters $\lbrace \widehat{\mathcal{C}}_k\rbrace_{k=1}^{K}$ of
    novel words and cluster $\mathcal{C}_{out}$ of outlier
    words. Rearrange the rows of $\widetilde{X}$ indexed by
    $\widehat{\mathcal{C}_k}$ into a matrix $Y_k$.
%
% subspace clustering
%
    \State For each $w \in \widehat{\mathcal{C}}_0\bigcup\mathcal{C}_{out}$,
    solve
   \begin{eqnarray*}
     \min\limits_{\lbrace{b}_{wl}\in \mathbb{R}_{+}^{\vert
         \widehat{\mathcal{C}}_l \vert}\rbrace_{l=1}^{K}} \Vert
     \widetilde{X}_w - \sum_{l=1}^K b_{wl} Y_l\Vert_2^2 + \lambda_2
     \sum_{l=1}^K \Vert b_{wl}\Vert_\infty
   \end{eqnarray*}
   for some $\lambda_2 \geq 0$. Let $\lbrace
   b_{wl}^{*}\rbrace_{l=1}^{K}$ be the optimal solution.
%
% mixture weight by Gourp Lasso
%
	\State For $w=1,\ldots,W$, $k=1,\ldots,K$, set
	\begin{equation*}
	\widehat{\beta}_{wk} =\left\{
        \begin{array}{lcl}
         N_w\one(w\in \widehat{\mathcal{C}}_k) & \mbox{for} &
         w\in\bigcup_{l=1}^{K}\widehat{\mathcal{C}}_l \\
         N_w \Vert b_{wk}^{*}\Vert_1 & \mbox{for} & w \in
         \widehat{\mathcal{C}}_0\bigcup\mathcal{C}_{out}
	\end{array}\right.
	\end{equation*}
	and normalize each column of $\widehat{\beta}$ to be column
        stochastic.
\end{algorithmic}
\end{algorithm}
%
%%%%%%%%%%%%%%%%%%%%%%%%%%%%%%%%%%%%%%%%%%%%%%%%%%%%%%%%%%%%%%%%%%%%%%%
%
% algorithm Random Projection
%
\vglue -2ex
\begin{algorithm}[!htb]
\caption{Find candidate novel words}
\label{randproj}
\begin{algorithmic}[1]
    \Require Set of $1\times M$ probability row-vectors
    $\widetilde{x}_1,\ldots,\widetilde{x}_W$; Number of projections
    $P$; Tolerance $\delta$.
    \Ensure Set $\mathcal{E}$ of candidate novel row-vectors.
    \State Set $\mathcal{E} = \emptyset$.
%
%  \State For $p = 1, \ldots, P$
%
     \State Generate row-vector $d\sim$ Uniform(unit-sphere in
     $\mathbb{R}^M)$.
        \State $i_{max} := \arg\max_{i} \widetilde{x}_i d^T$, $i_{min}
        := \arg\min_{i} \widetilde{x}_i d^T$.
%
%. Add $x_{max}, x_{min}$ to extreme point sets $Y_s$.
%
        \State $\mathcal{E} \leftarrow \mathcal{E} \bigcup \lbrace
        x_i: \Vert x_i - x_{i_{max}} \Vert_1 \leq \delta \mbox{ or }
        \Vert x_i - x_{i_{min}} \Vert_1 \leq \delta \rbrace$.
        \label{step5}
        \State Repeat steps $2$ through $4$, $P$ times.
\end{algorithmic}
\end{algorithm}
\vglue -2ex
Step (2) of Algorithm~\ref{alg1} finds rows of $\tilde{X}$ many of
which are likely to correspond to the novel words of topics and some
to outliers (non-novel words). This step uses Algorithm~\ref{randproj}
which is a linear-complexity procedure for finding, with high
probability, extreme points and points close to them (the candidate
novel words of topics) using a small number $P$ of random
projections. Step (3) uses the state-of-the-art sparse subspace
clustering algorithm from \cite{SSC_candies:ref,SSC_vidal:ref} to
identify $K$ clusters of novel words, one for each topic, and an
additional cluster containing the outliers (non-novel words). Step (4)
expresses rows of $\widetilde{X}$ corresponding to non-novel words as
convex combinations of these $K$ groups of rows and step (5) estimates
the entries in the topic matrix and normalizes it to make it
column-stochastic.
%
% the Group sparsity issue
%
In many applications, non-novel words occur in only a few topics.  The
{\it group-sparsity} penalty $\lambda_2 \sum_{l=1}^K \|
b_{wl}\|_\infty$ proposed in \cite{convexNMF:ref} is used in step (4)
of Algorithm~\ref{alg1} to favor solutions where the row vectors of
non-novel words are convex combinations of as few groups of novel
words as possible.
%
% Polynomial-time solution
%
Our proposed algorithm runs in polynomial-time in $W$, $M$, and $K$
and all the optimization problems involved are convex.

%%%%%%%%%%%%%%%%%%%%%%%%%%%%%%%%%%%%%%%%%%%%%%%%%%%%%%%%%%%%%%%%%%%%%%%%%%%%
%%
%% algorithm SSC
%%
%\begin{algorithm}[!htb]
%%
%\caption{Sparse Subspace Clustering}
%\label{SSC}
%%
%\begin{algorithmic}[1]
%%
%    \Require Set of $r$ row-vectors $\mathcal{E}=\lbrace
%    y_1,\ldots,y_r\rbrace$, arranged as $Y\in\mathbb{R}^{r\times M} $;
%    Number of clusters $K$; Threshold $\gamma \leq 1$
%%
%   \Ensure A clustering $\lbrace \mathcal{C}_k\rbrace_{k=1}^{K}$ of
%   $\mathcal{E}$ and a outlier set $\mathcal{C}_{out}$.
%%
%    \State For $i=1,\ldots,r$,for some $\lambda_1 >0$, find solution
%    $c_i^*$ to
%%
%%\begin{equation}
%%\label{SSC-equation}
%%\min\limits_{\vec{c}_i} \Vert \vec{c}_i\Vert_{\ell_1},  subject \quad to\quad \vec{y}_i=\vec{c} Y,  and \quad [\vec{c}_i]_i=0
%%\end{equation}
%%or
%%
%	\begin{eqnarray*}
%%
%	\label{Lasso-equation}
%%
%	& & \min\limits_{c_i}   \lambda_1 ||c_i||_1 + \Vert y_i-c_i Y \Vert_2^2 \\
%	& & s.t. \quad [c_i]_i=0, c_i \in \mathbb{R}^r
%%
%	\end{eqnarray*}
%%
%	If optimal cost $\Vert c_i\Vert_1 \leq \gamma$, then claim
%        $y_i \in \mathcal{C}_{out}$.
%%
%    \State Set $C=[{c}_1^{*T},\ldots,{c}_{r'}^{*T}]$ without
%    outliers. Set $E = |C| + |C|^T$. Form graph $\mathcal{G}$ with
%    nodes representing $r'$ vectors and edge weights given by $E$.
%%
%    \State Apply spectral clustering algorithm to $G$ with number of
%    clusters $K$. Obtain a clustering $\lbrace
%    \mathcal{C}_k\rbrace_{k=1}^{K}$.
%%
%\end{algorithmic}
%%
%\end{algorithm}
%%
%%%%%%%%%%%%%%%%%%%%%%%%%%%%%%%%%%%%%%%%%%%%%%%%%%%%%%%%%%%%%%%%%%%%%%%%%%%%%

\section{Experimental results}
\subsection{Synthetic Dataset}
\label{sec:synthetic}
\vglue -4ex
\begin{figure}[htb]
\begin{minipage}[b]{.90\linewidth}
  \centering
  \centerline{\includegraphics[width=8.0cm]{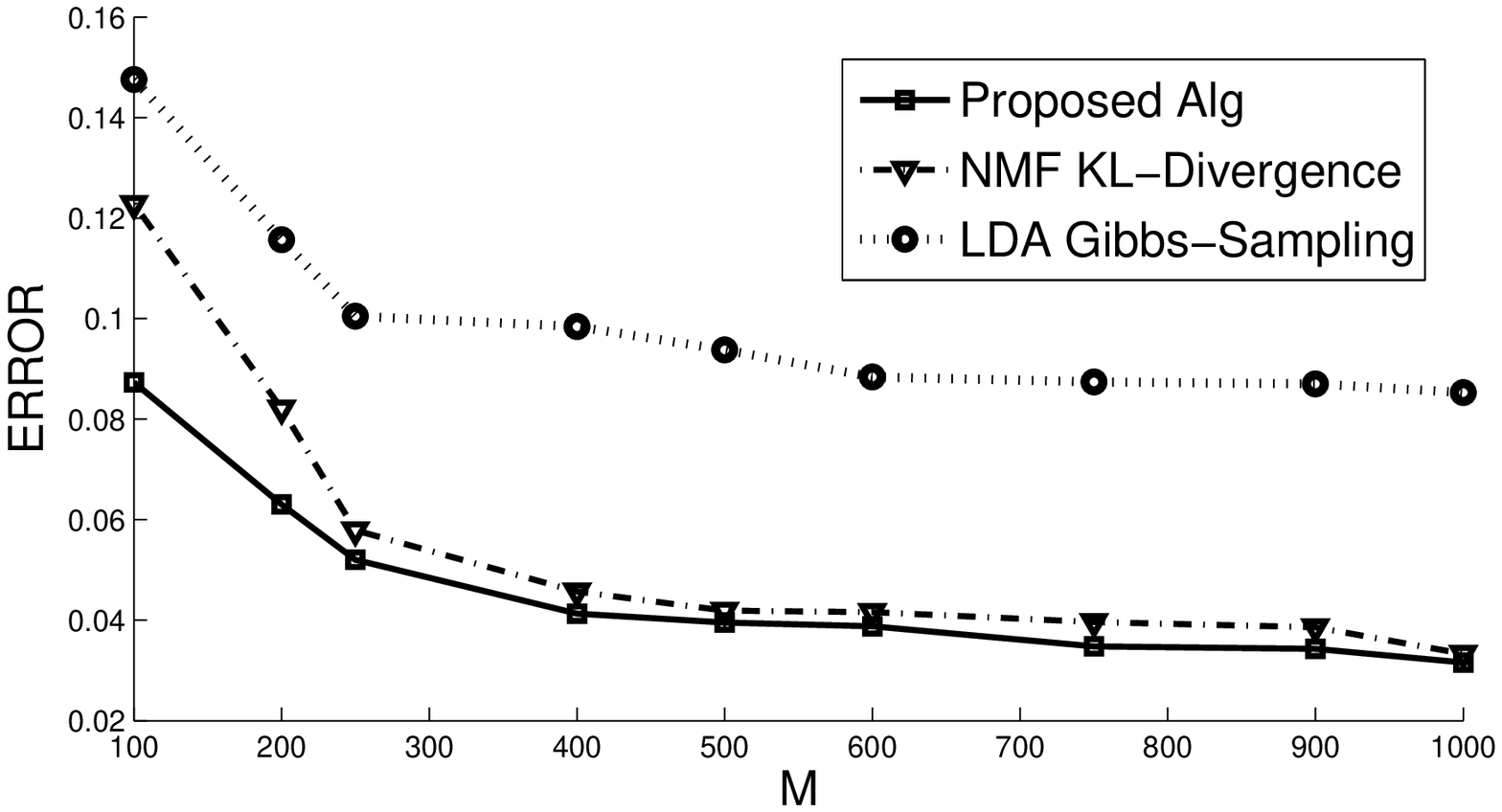}}
%  \vspace{1.5cm}
%  \centerline{(a)}\medskip
\end{minipage}
\vfill
\begin{minipage}[b]{0.90\linewidth}
  \centering
  \centerline{\includegraphics[width=8.0cm]{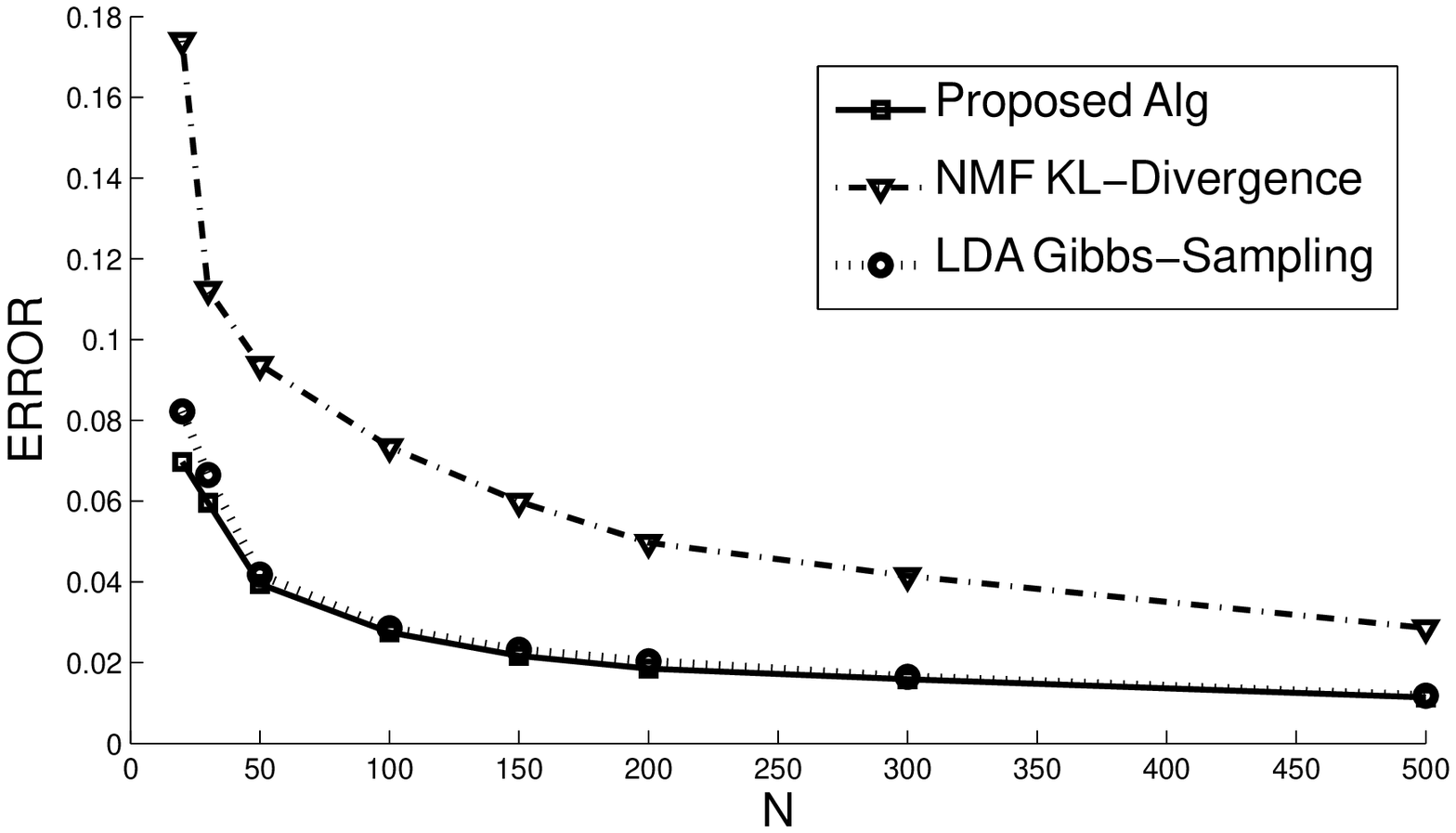}}
%  \vspace{1.5cm}
%  \centerline{(b)}\medskip
\end{minipage}
\caption{Error of estimated topic matrix in Frobenius norm. Upper: $W
  = 500,\rho = 0.2, N = 50, K = 5$; Lower: $W = 500, \rho = 0.2, K =
  10, M = 500$.}
\label{fig:synthetic}
\end{figure}
\vglue -3ex
In this section, we validate our algorithm on some synthetic examples.
% Generate topic matrix
We generate a $W \times K$ separable topic matrix $\beta$ with $W_1/K
> 1$ novel words per topic as follows: first, iid $1\times K$
rows-vectors corresponding to non-novel words are generated uniformly
on the probability simplex. Then, $W_1$ iid $\mathrm{Uniform}[0,1]$
values are generated for the nonzero entries in the rows of novel
words. The resulting matrix is then column-normalized to get one
realization of $\beta$. Let $\rho := W_1/W$. Next, $M$ iid $K\times 1$
column-vectors are generated for the $\theta$ matrix according to a
Dirichlet prior $c\prod\limits_{i=1}^{K} \theta_i^{\alpha_i
  -1}$. Following \cite{Griffiths:ref}, we set $\alpha_i = 0.1$ for
all $i$. Finally, we obtain $X$ by generating $N$ iid words for each
document.

For different settings of $W$, $\rho$, $K$, $M$ and $N$, we calculate
the error of the estimated topic matrix $\widehat{\beta}$ as $\Vert
\widehat{\beta}-\beta \Vert_{F}$. For each setting we average the
error over $50$ random samples. In sparse subspace clustering
%Algorithm~\ref{SSC}),
the value of $\lambda_1$ is set as in \cite{SSC_vidal:ref} (it depends
on the size of the candidate set) and the value of $\gamma$ as in
\cite{SSC_candies:ref} (it depends on the values of $N,M$). In Step 4
of Algorithm~\ref{alg1}, we set $\lambda_2 = 0.01$
%to a small constant 
for all settings.
%as we have no prior knowledge about the part of topic matrix for
%non-novel words.

We compare our algorithm against the LDA algorithm \cite{LDA:ref} and
a state-of-art NMF-based algorithm \cite{betaDivergence:ref}. This NMF
algorithm is chosen because it compensates for the type of noise we
use in our topic model. Our LDA algorithm uses Gibbs sampling for
inferencing. Figure~\ref{fig:synthetic} depicts the estimation error
as a function of the number of documents $M$ (top) and the number of
words/document $N$ (bottom). Evidently, our algorithm is uniformly
better than comparable techniques. Specifically, while NMF has similar
error as our algorithm for large $M$ it performs relatively poorly as
a function of $N$. On the other hand LDA has similar error performance
as ours for large $N$ but performs poorly as a function of $M$. Note
that both of these algorithms have comparably high error rates for
small $M$ and $N$.
%
%is depicted at the top of Fig.~\ref{fig:synthetic} and 
%In the upper plot we show the
%estimation error as number of document $M$ increase while in the lower
%we show the effect of ``words'' per document $N$. As shown in figure
%\ref{fig:synthetic}, our algorithm performs better in all these
%settings especially when number of samples are small, either in terms
%of number of documents or number of words.
\vspace{-2ex}

\subsection{Swimmer Image Dataset}
\vglue -2ex
\begin{figure}[!htb]
\begin{minipage}[b]{.9\linewidth}
\centering
\begin{minipage}[b]{1.0cm}
\centerline{
\includegraphics[width=1.0cm]{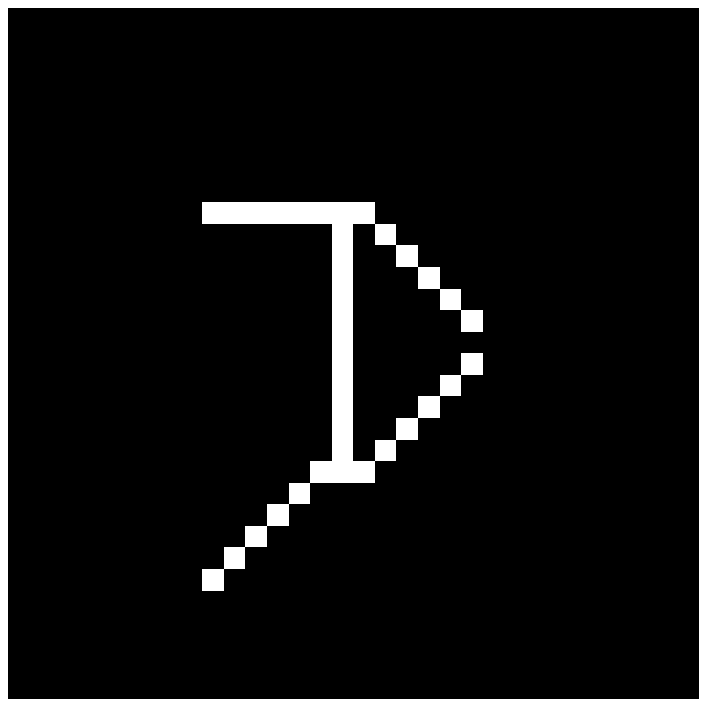}}
\end{minipage}
\begin{minipage}[b]{1.0cm}
\centerline{
\includegraphics[width=1.0cm]{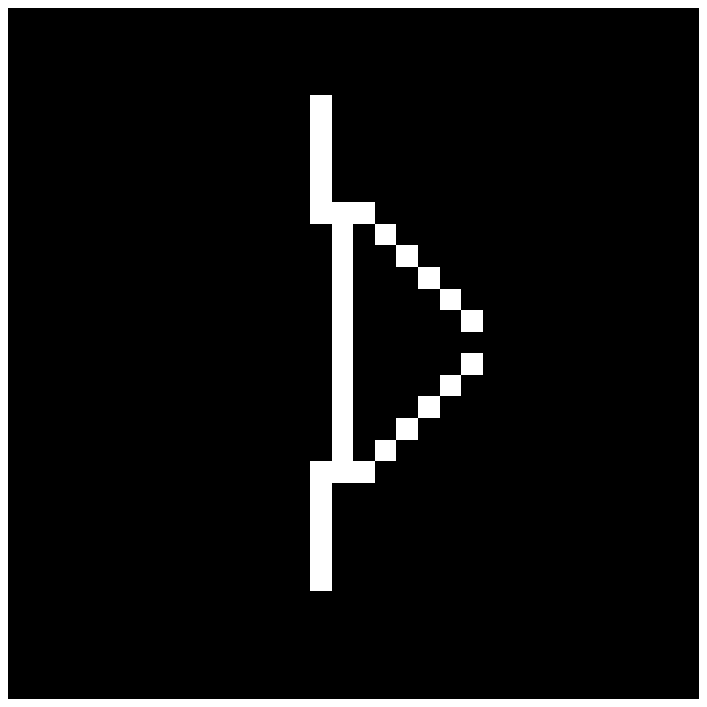}}
\end{minipage}
\begin{minipage}[b]{1.0cm}
\centerline{
\includegraphics[width=1.0cm]{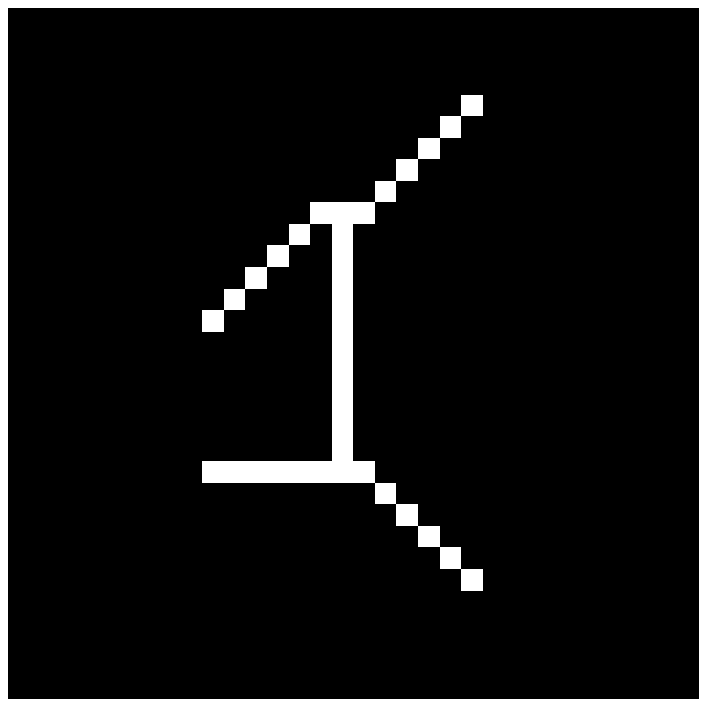}}
\end{minipage}
\begin{minipage}[b]{1.0cm}
\centerline{
\includegraphics[width=1.0cm]{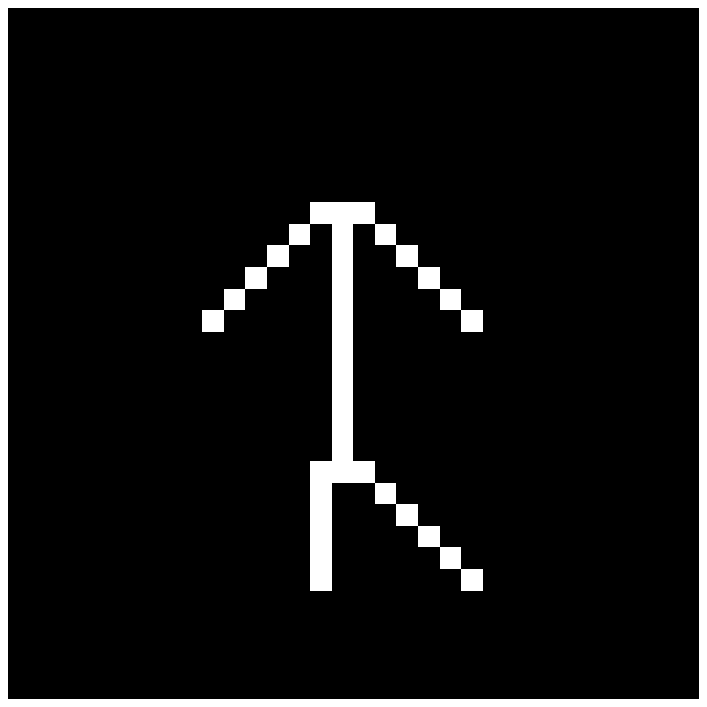}}
\end{minipage}
(a)
\end{minipage}
\begin{minipage}[b]{.9\linewidth}
\centering
\begin{minipage}[b]{1.0cm}
\centerline{
\includegraphics[width=1.0cm]{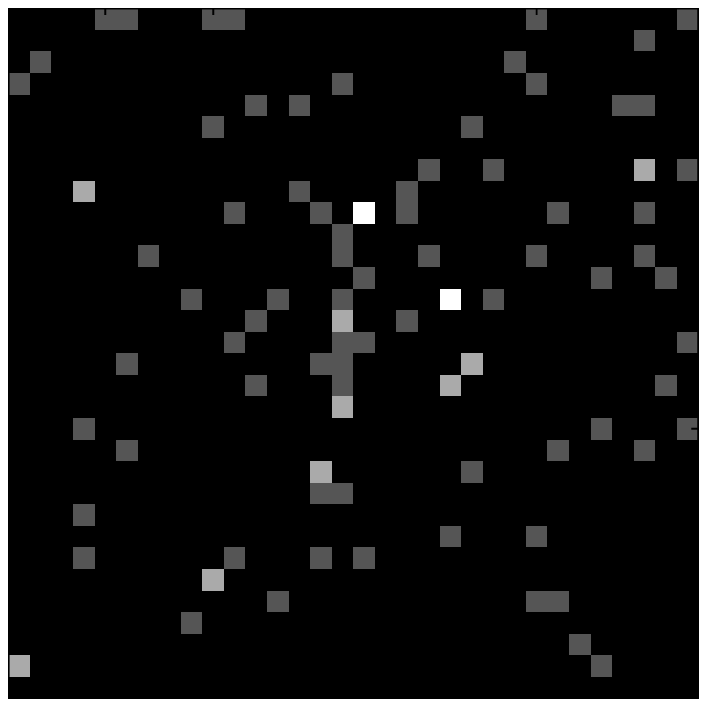}}
\end{minipage}
\begin{minipage}[b]{1.0cm}
\centerline{
\includegraphics[width=1.0cm]{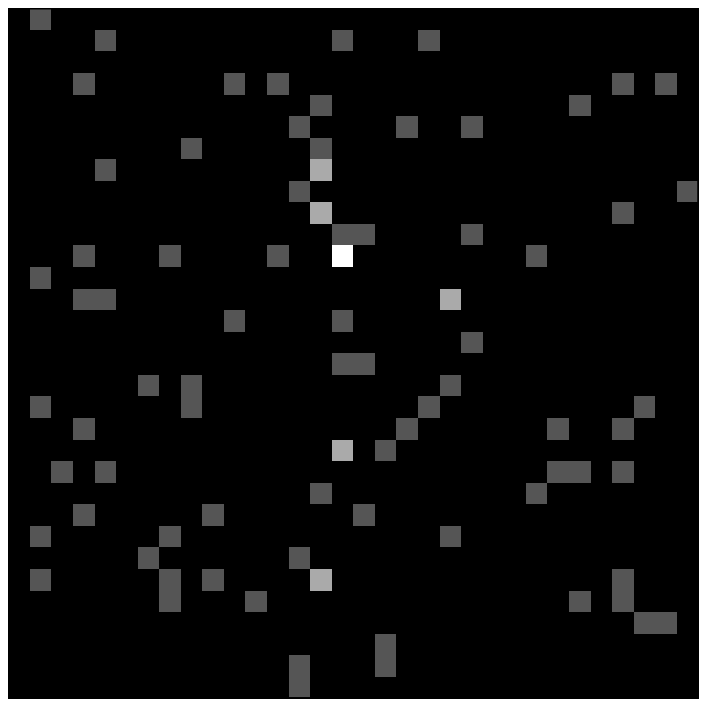}}
\end{minipage}
\begin{minipage}[b]{1.0cm}
\centerline{
\includegraphics[width=1.0cm]{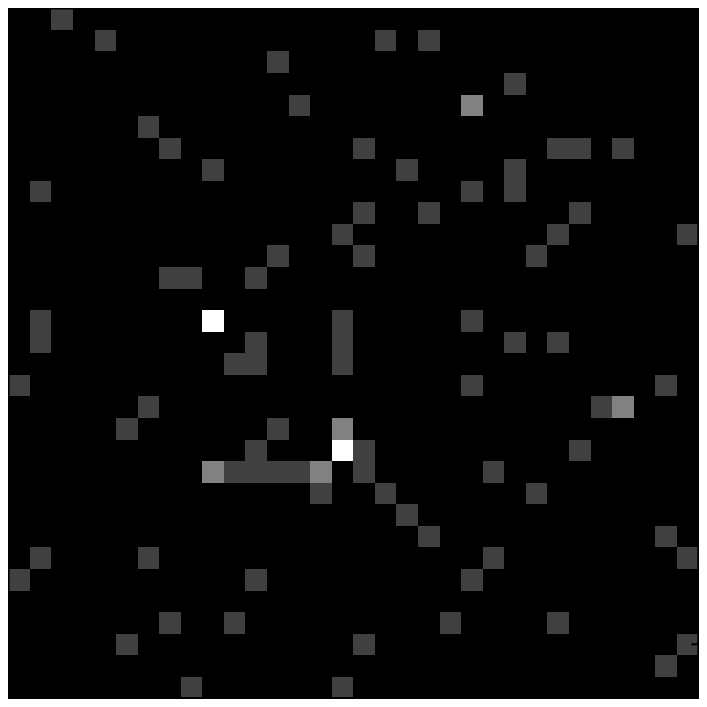}}
\end{minipage}
\begin{minipage}[b]{1.0cm}
\centerline{
\includegraphics[width=1.0cm]{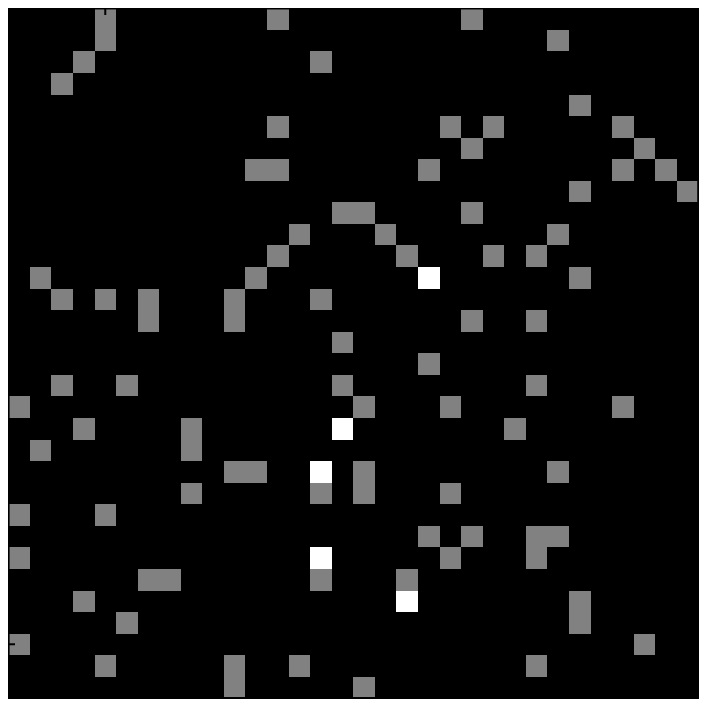}}
\end{minipage}
(b)
\end{minipage}
\begin{minipage}[b]{.9\linewidth}
\centering
\begin{minipage}[b]{1.0cm}
\centerline{
\includegraphics[width=1.0cm]{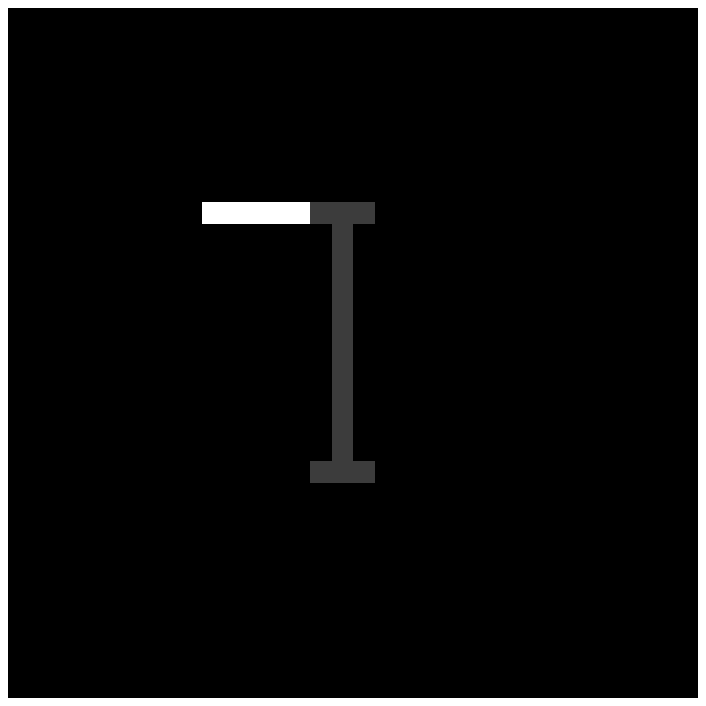}}
\end{minipage}
\begin{minipage}[b]{1.0cm}
\centerline{
\includegraphics[width=1.0cm]{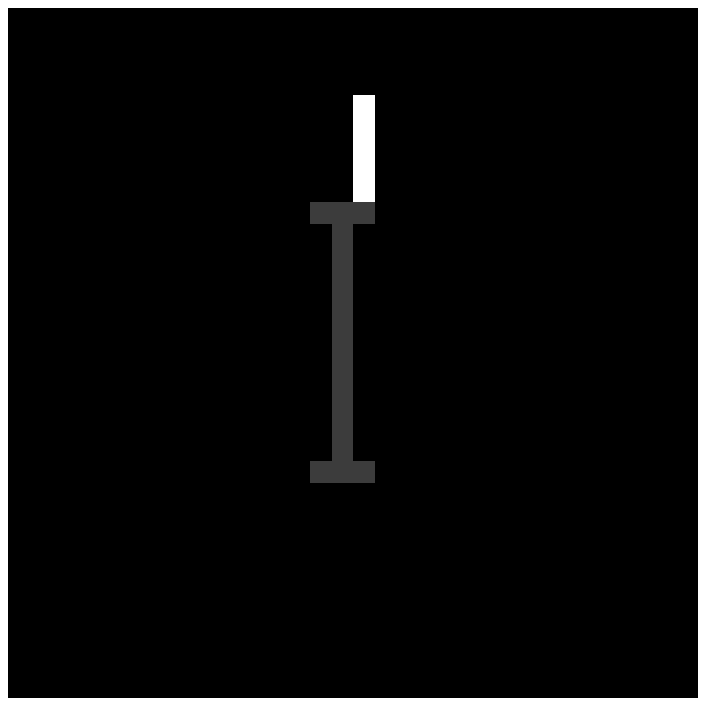}}
\end{minipage}
\begin{minipage}[b]{1.0cm}
\centerline{
\includegraphics[width=1.0cm]{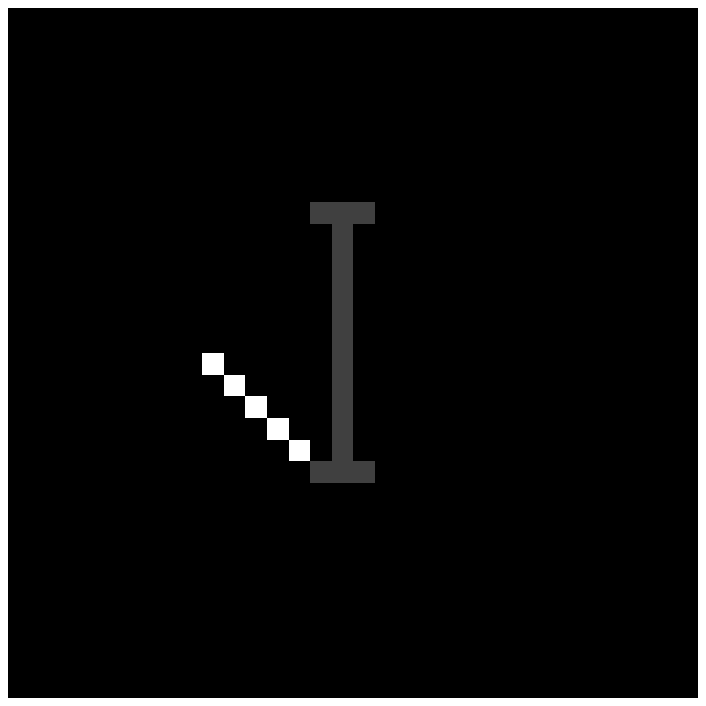}}
\end{minipage}
\begin{minipage}[b]{1.0cm}
\centerline{
\includegraphics[width=1.0cm]{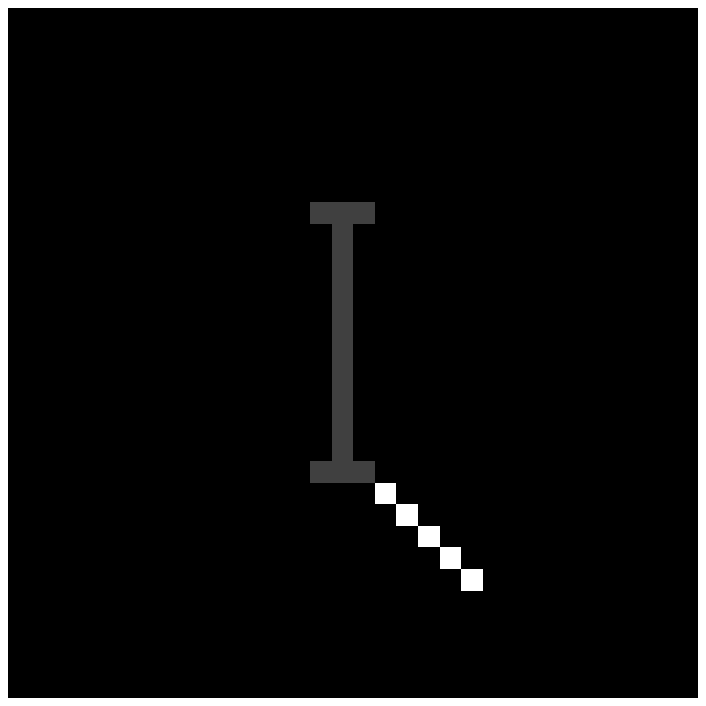}}
\end{minipage}
(c)
\end{minipage}
\vglue -1ex
\caption{(a) Example ``clean'' images (cols.~of $A$) in Swimmer
  dataset; (b) Corresponding images with sampling ``noise'' (cols.~of
  $X$); (c) Examples of ideal topics (cols.~of $\beta$).}
\label{fig:sample_swimmer}
\end{figure}
%

%%%%%%%%%%%%%%%%%%%%%%%%%%%%%%%%%
\begin{figure*}[!htb]
\begin{tabular}{|m{0.3cm}|@{\hskip 0.05cm}m{1.0cm}@{\hskip 0.05cm}m{1.0cm}@{\hskip 0.05cm}m{1.0cm}@{\hskip 0.05cm}m{1.0cm}@{\hskip 0.05cm}m{1.0cm}@{\hskip 0.05cm}m{1.0cm}@{\hskip 0.05cm}m{1.0cm}@{\hskip 0.05cm}m{1.0cm}@{\hskip 0.05cm}m{1.0cm}@{\hskip 0.05cm}m{1.0cm}@{\hskip 0.05cm}m{1.0cm}@{\hskip 0.05cm}m{1.0cm}@{\hskip 0.05cm}m{1.0cm}@{\hskip 0.05cm}m{1.0cm}@{\hskip 0.05cm}m{1.0cm}@{\hskip 0.05cm}m{1.0cm}@{\hskip 0.05cm}|}
%\
\hline
Pos. & LA 1 & LA 2& LA 3& LA 4& RA 1 & RA 2& RA 3& RA 4& LL 1 & LL 2&
LL 3& LL 4& RL 1 & RL 2& RL 3& RL 4 \\
\hline
a)
&
\parbox[c]{1cm}{\vspace*{0.1cm}\includegraphics[width=1.0cm]{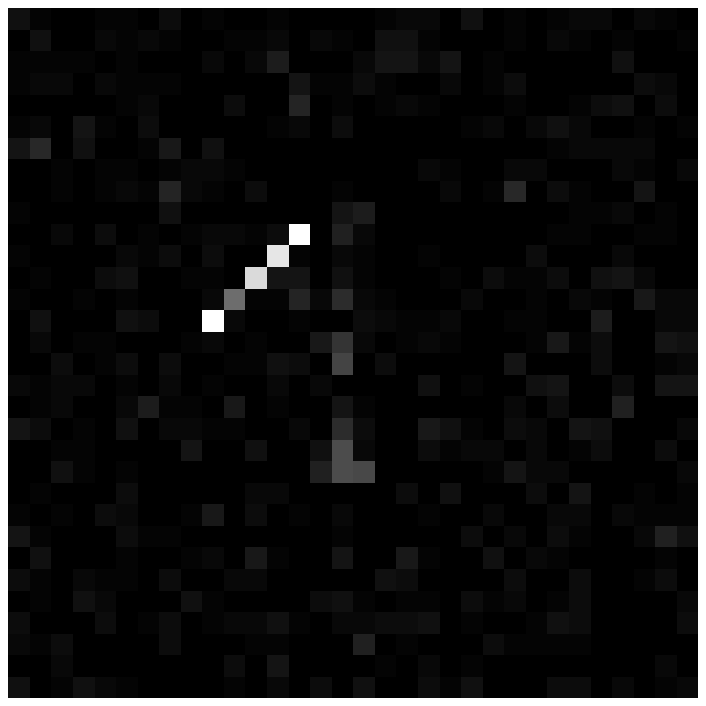}
\vspace*{-0.3cm}}
&
\parbox[c]{1cm}{\vspace*{0.1cm}\includegraphics[width=1.0cm]{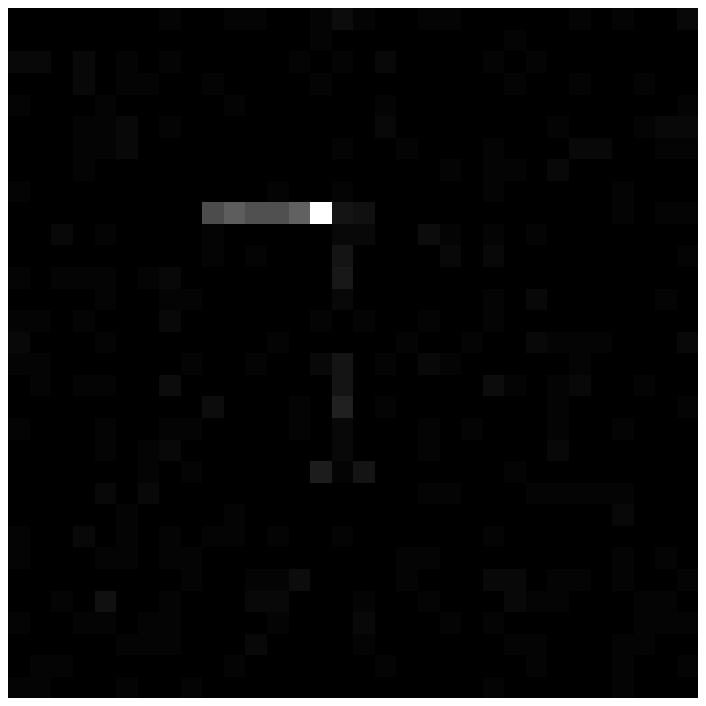}
\vspace*{-0.3cm}}
&
\parbox[c]{1cm}{\vspace*{0.1cm}\includegraphics[width=1.0cm]{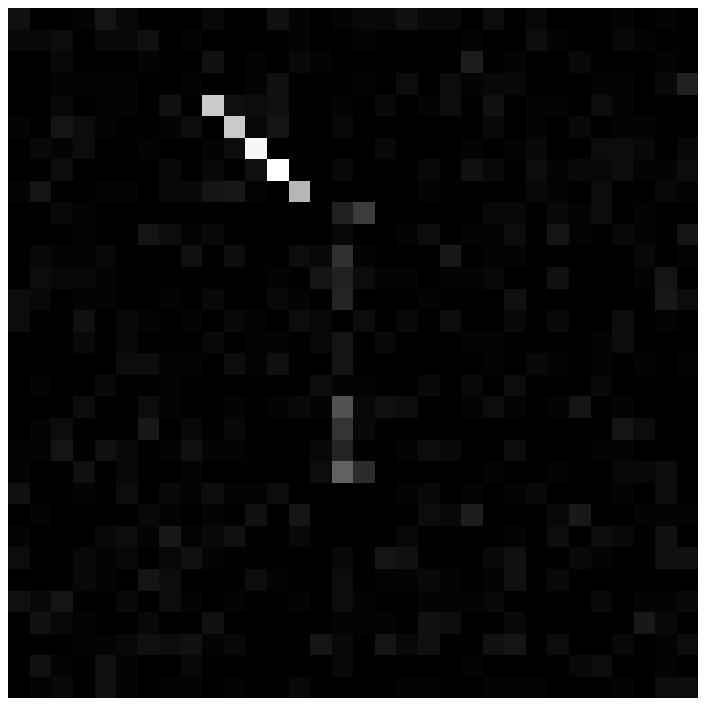}
\vspace*{-0.3cm}}
&
\parbox[c]{1cm}{\vspace*{0.1cm}\includegraphics[width=1.0cm]{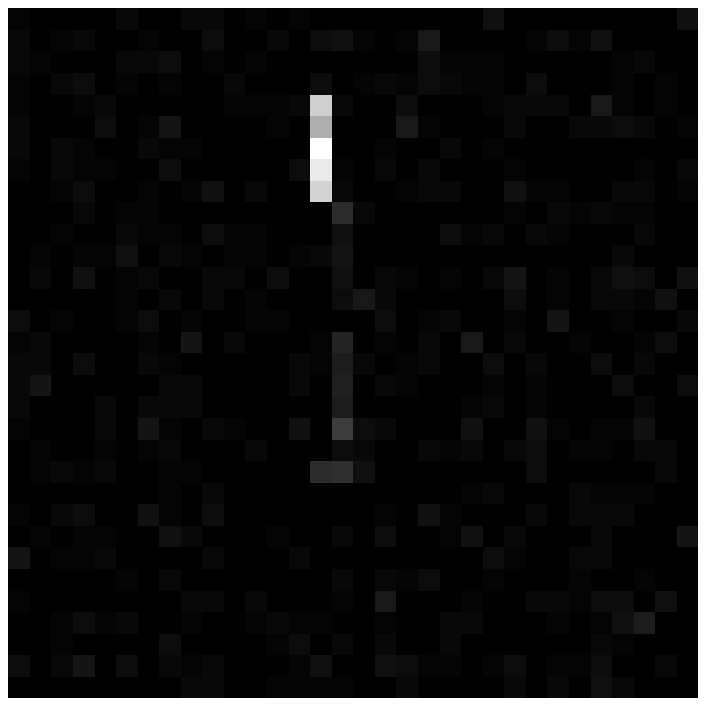}
\vspace*{-0.3cm}}
&
\parbox[c]{1cm}{\vspace*{0.1cm}\includegraphics[width=1.0cm]{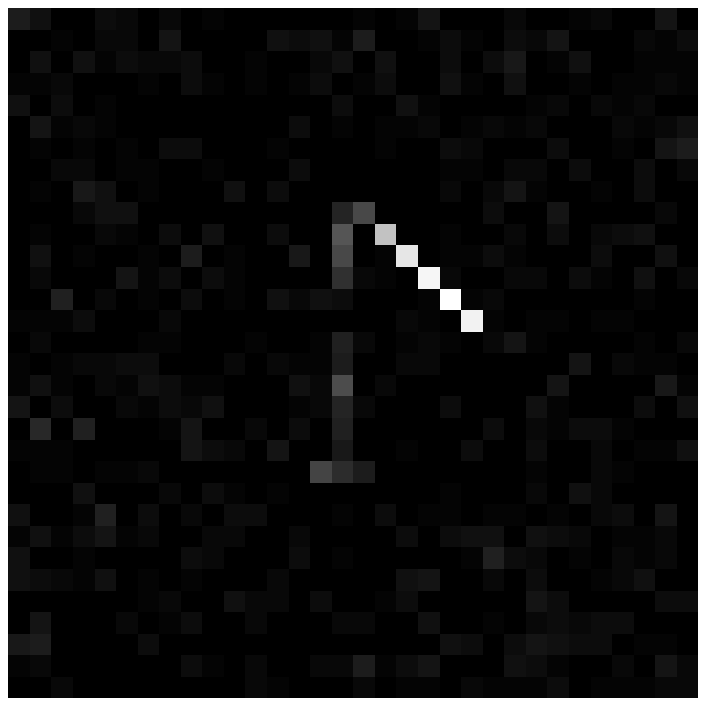}
\vspace*{-0.3cm}}
&
\parbox[c]{1cm}{\vspace*{0.1cm}\includegraphics[width=1.0cm]{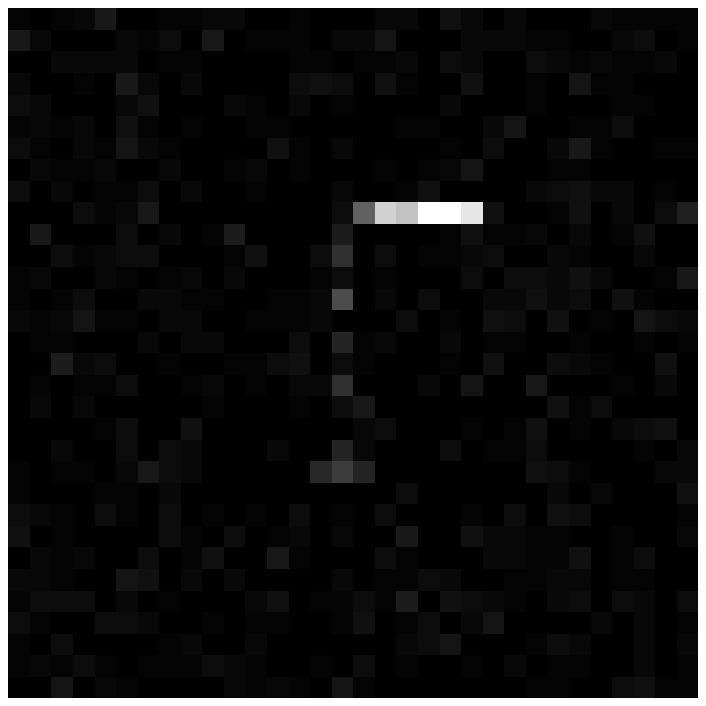}
\vspace*{-0.3cm}}
&
\parbox[c]{1cm}{\vspace*{0.1cm}\includegraphics[width=1.0cm]{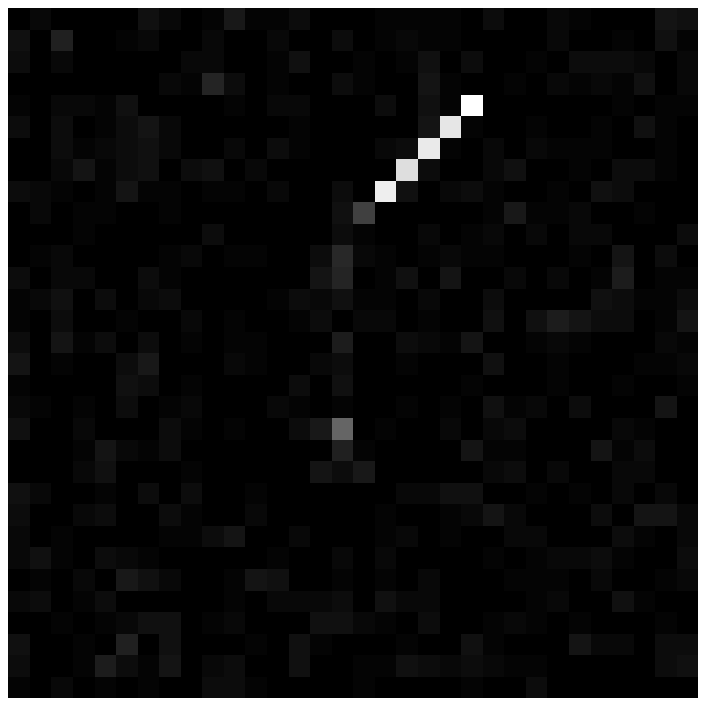}
\vspace*{-0.3cm}}
&
\parbox[c]{1cm}{\vspace*{0.1cm}\includegraphics[width=1.0cm]{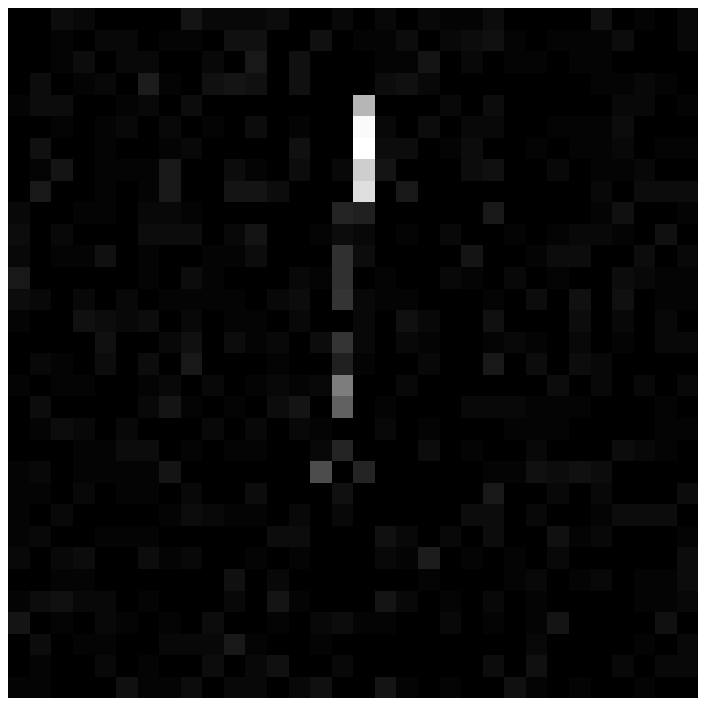}
\vspace*{-0.3cm}}
&
\parbox[c]{1cm}{\vspace*{0.1cm}\includegraphics[width=1.0cm]{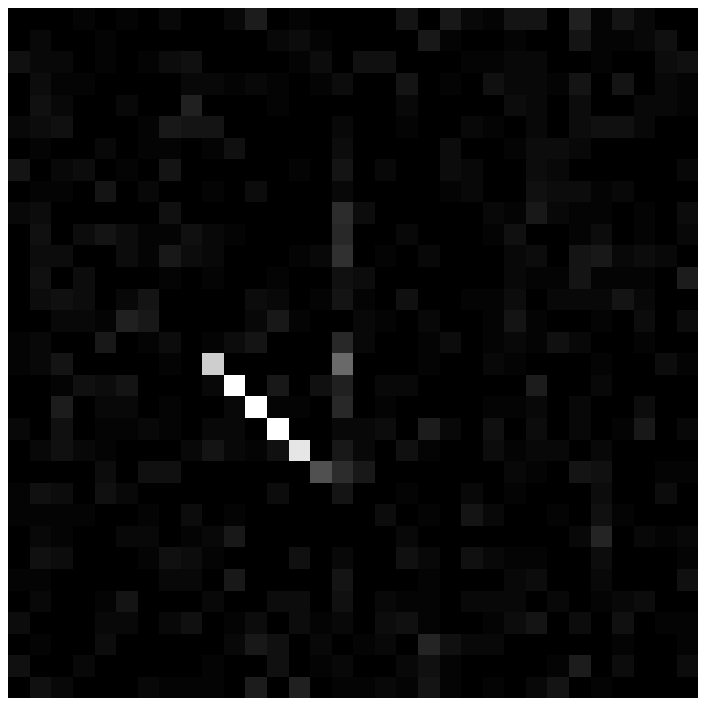}
\vspace*{-0.3cm}}
&
\parbox[c]{1cm}{\vspace*{0.1cm}\includegraphics[width=1.0cm]{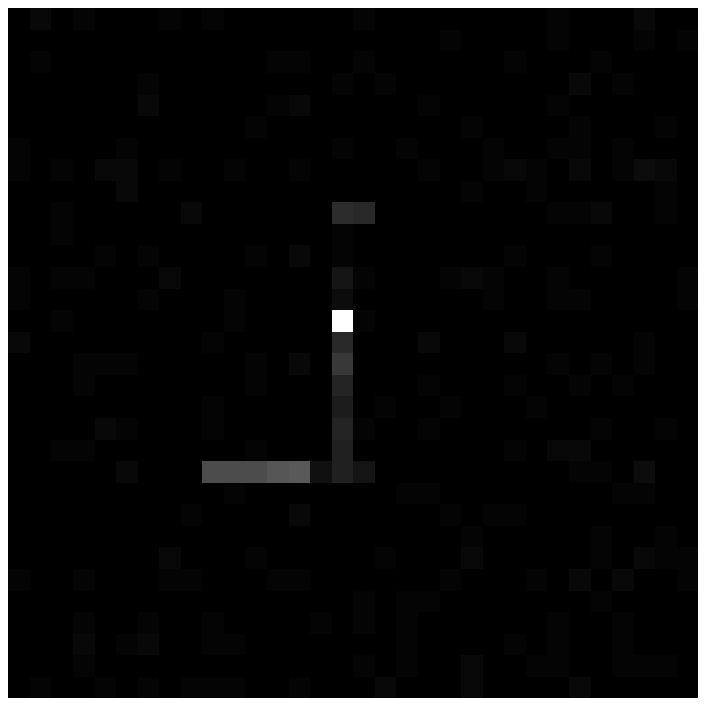}
\vspace*{-0.3cm}}
&
\parbox[c]{1cm}{\vspace*{0.1cm}\includegraphics[width=1.0cm]{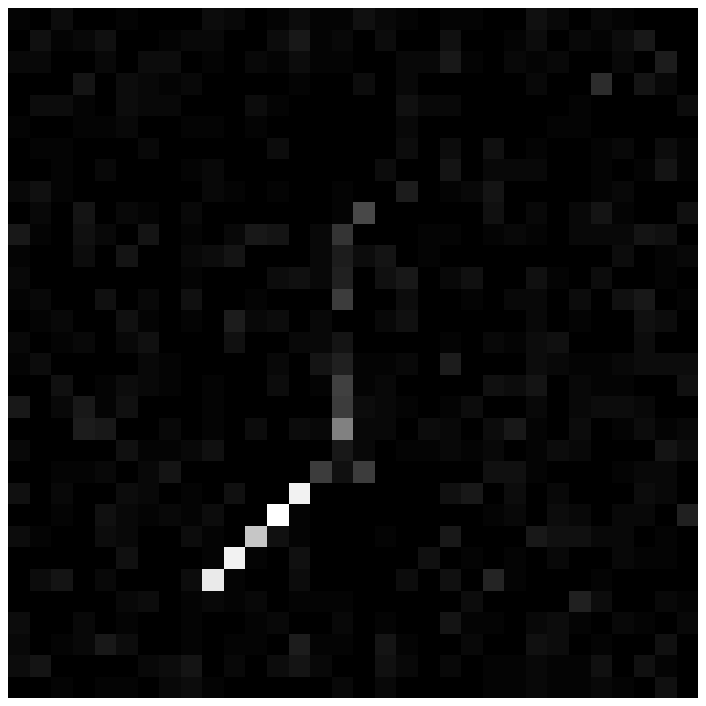}
\vspace*{-0.3cm}}
&
\parbox[c]{1cm}{\vspace*{0.1cm}\includegraphics[width=1.0cm]{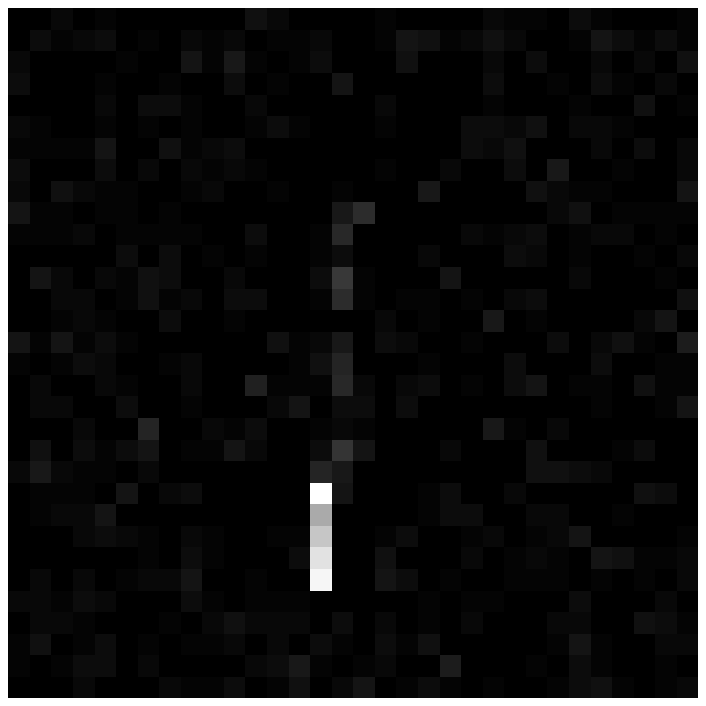}
\vspace*{-0.3cm}}
&
\parbox[c]{1cm}{\vspace*{0.1cm}\includegraphics[width=1.0cm]{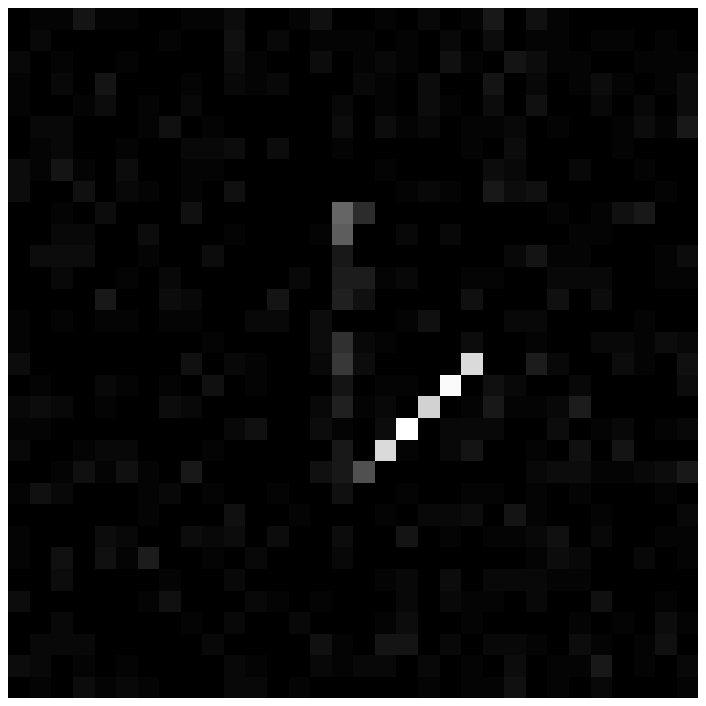}
\vspace*{-0.3cm}}
&
\parbox[c]{1cm}{\vspace*{0.1cm}\includegraphics[width=1.0cm]{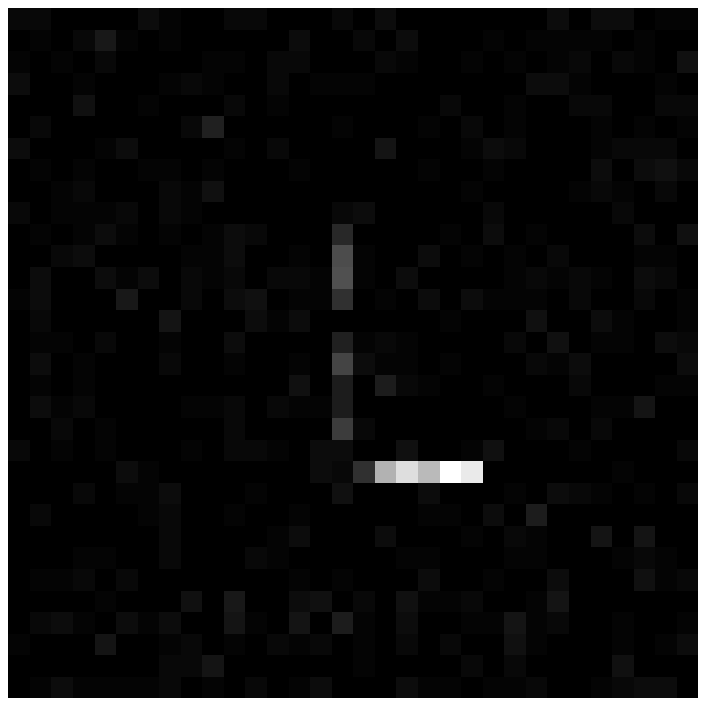}
\vspace*{-0.3cm}}
&
\parbox[c]{1cm}{\vspace*{0.1cm}\includegraphics[width=1.0cm]{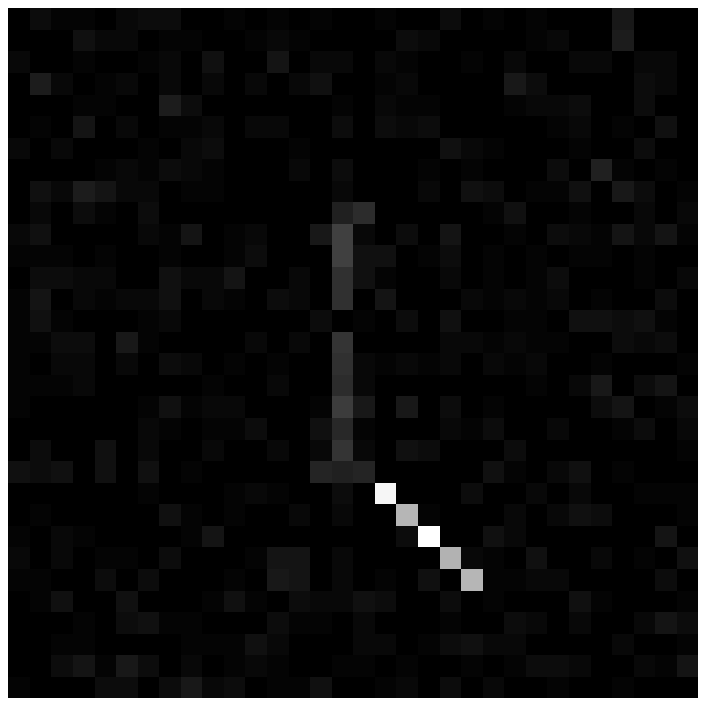}
\vspace*{-0.3cm}}
&
\parbox[c]{1cm}{\vspace*{0.1cm}\includegraphics[width=1.0cm]{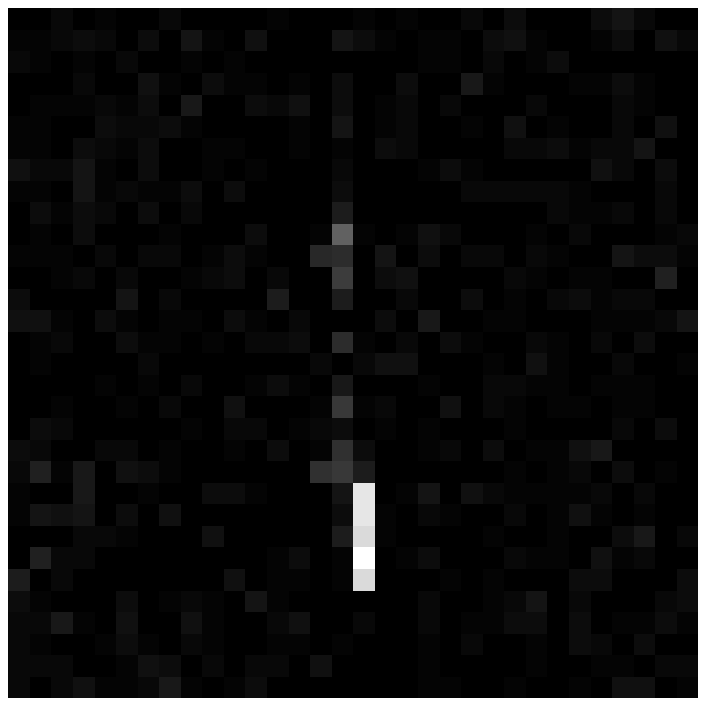}
\vspace*{-0.3cm}}
 \\%[0.5cm]
\hline
b) &
\parbox[c]{1cm}{\vspace*{0.1cm}\includegraphics[width=1.0cm]{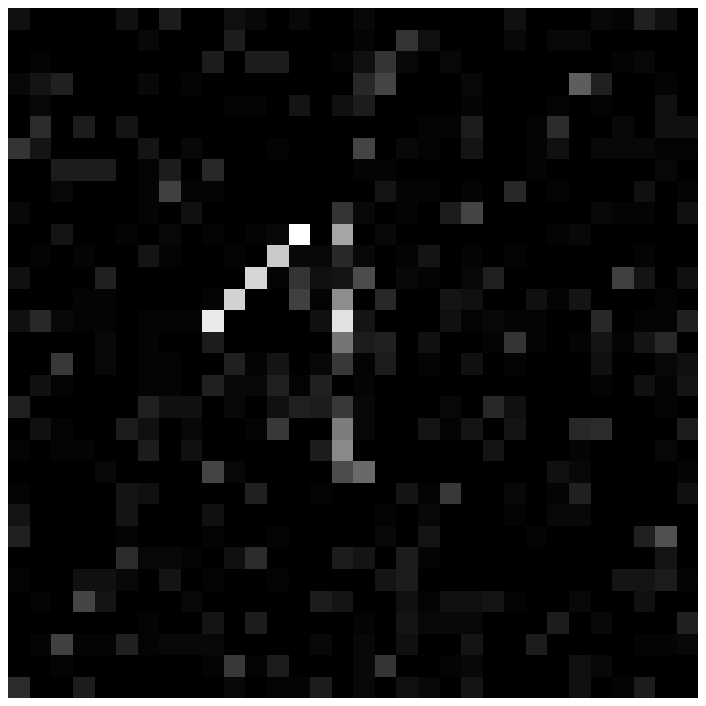}
\vspace*{-0.3cm}}
&
\parbox[c]{1cm}{\vspace*{0.1cm}\includegraphics[width=1.0cm]{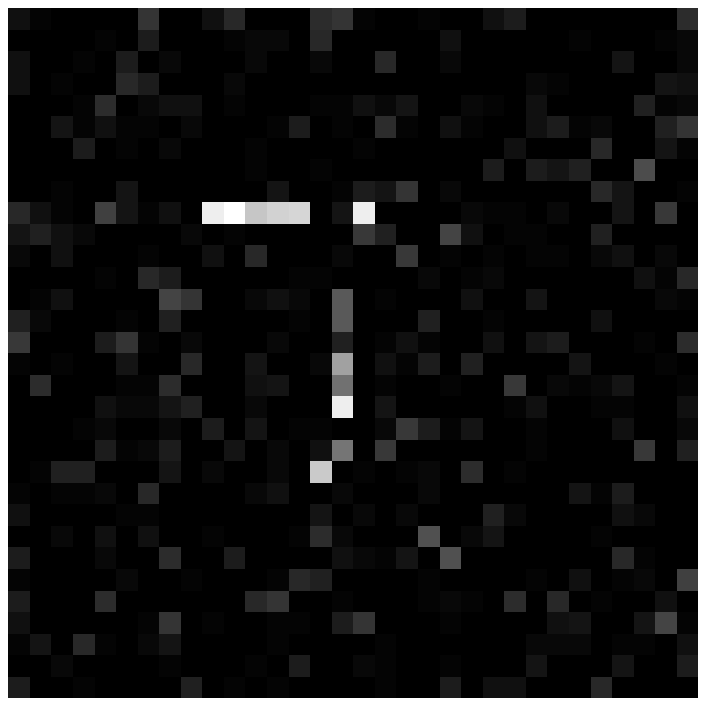}
\vspace*{-0.3cm}}
&
\parbox[c]{1cm}{\vspace*{0.1cm}\includegraphics[width=1.0cm]{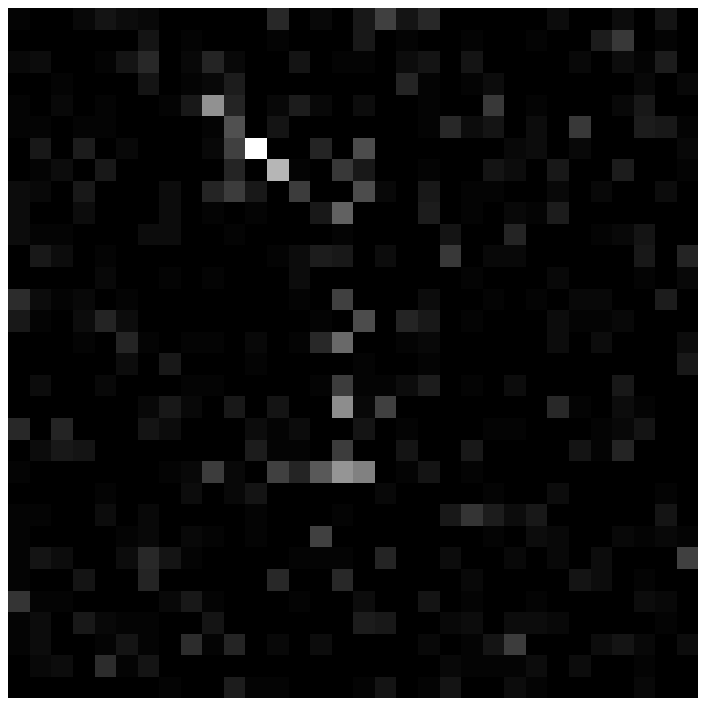}
\vspace*{-0.3cm}}
&
&
\parbox[c]{1cm}{\vspace*{0.1cm}\includegraphics[width=1.0cm]{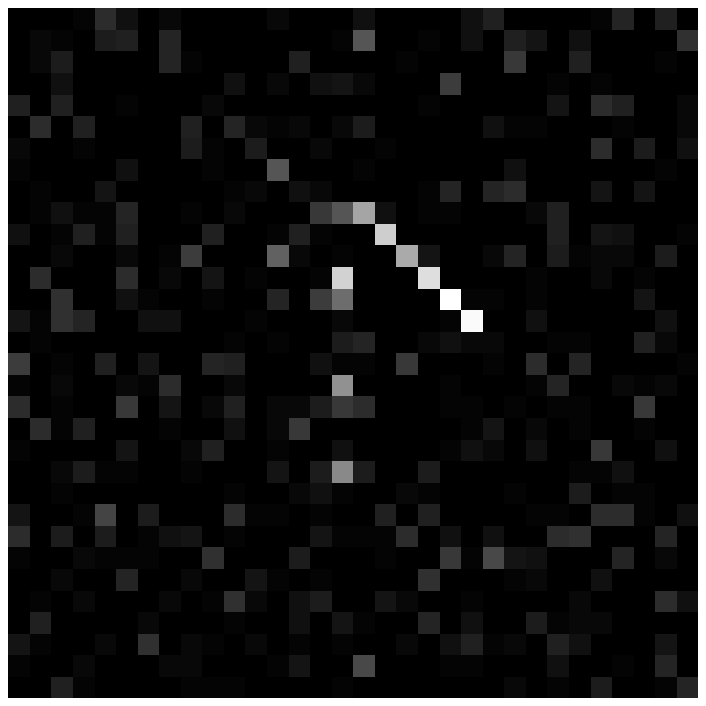}
\vspace*{-0.3cm}}
&
&
\parbox[c]{1cm}{\vspace*{0.1cm}\includegraphics[width=1.0cm]{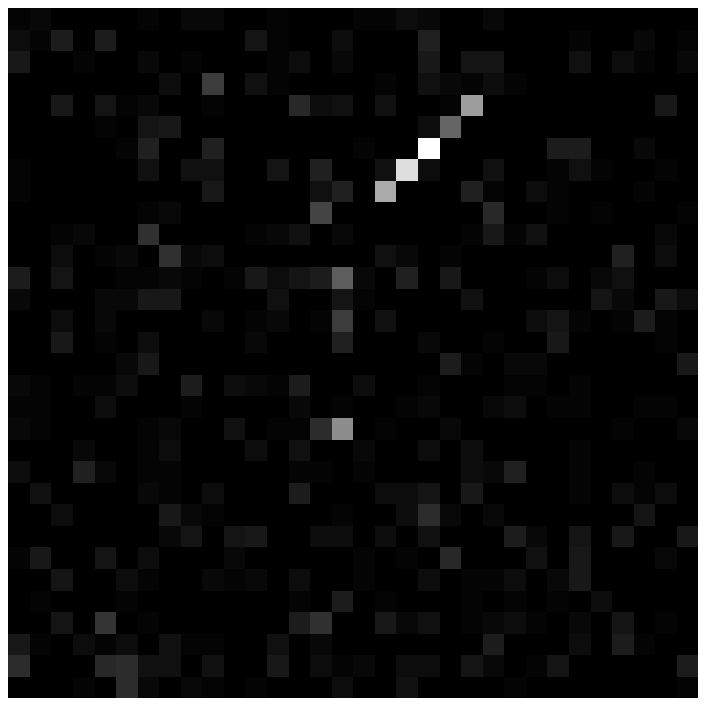}
\vspace*{-0.3cm}}
&
\parbox[c]{1cm}{\vspace*{0.1cm}\includegraphics[width=1.0cm]{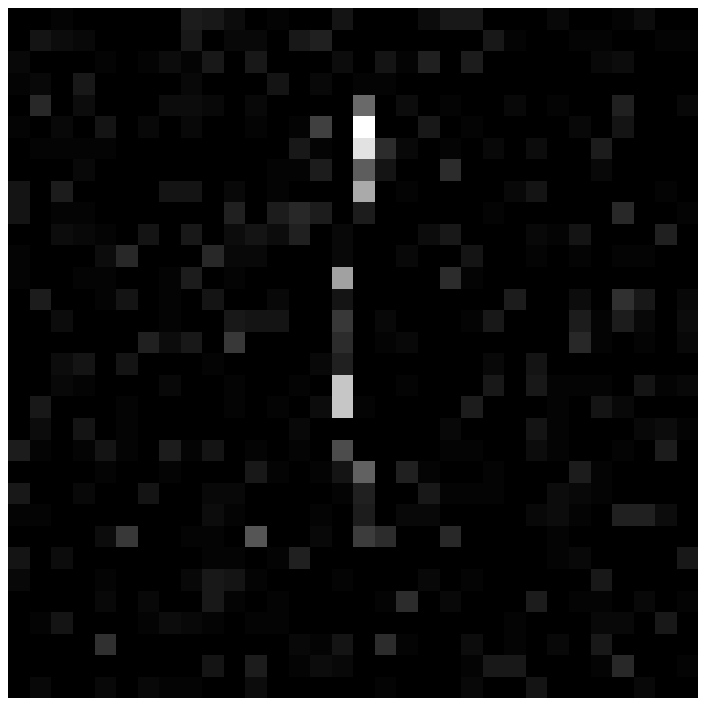}
\vspace*{-0.3cm}}
&
\parbox[c]{1cm}{\vspace*{0.1cm}\includegraphics[width=1.0cm]{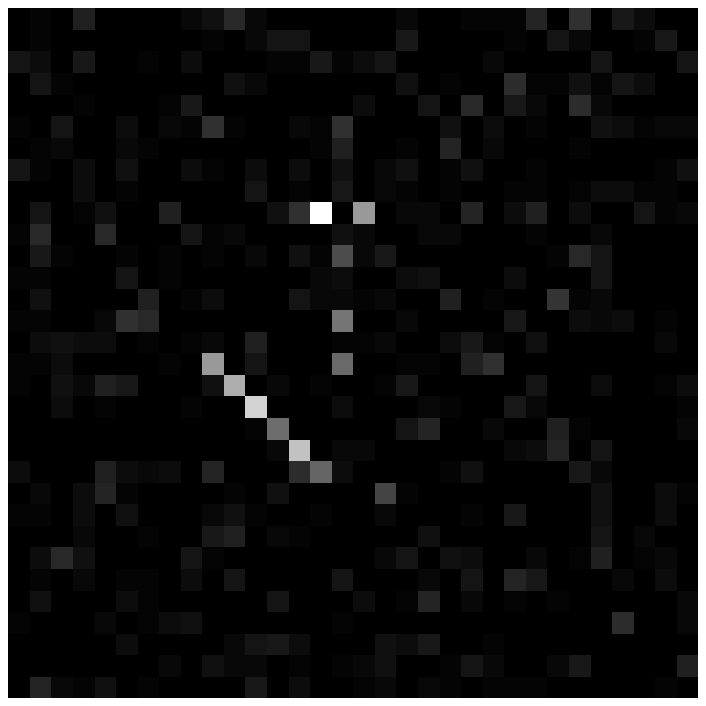}
\vspace*{-0.3cm}}
&
&
&
\parbox[c]{1cm}{\vspace*{0.1cm}\includegraphics[width=1.0cm]{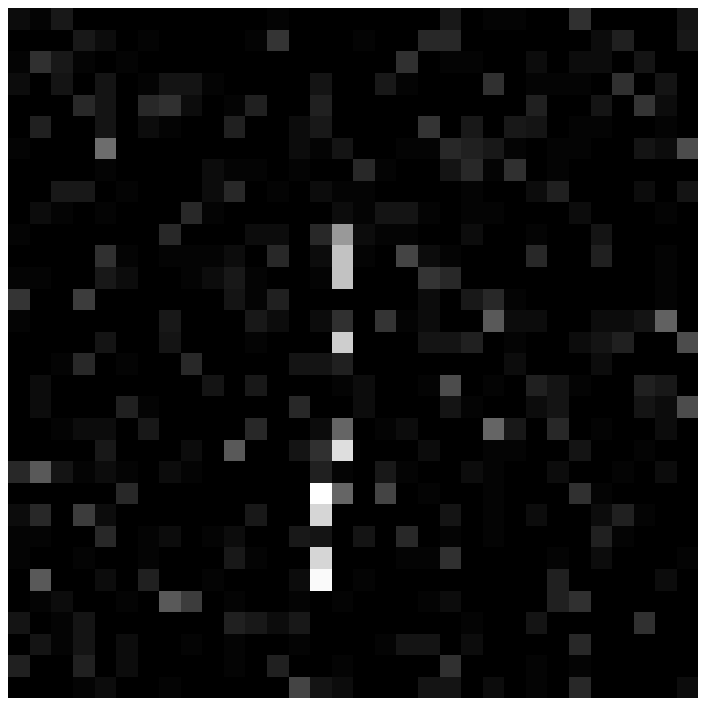}
\vspace*{-0.3cm}}
&
\parbox[c]{1cm}{\vspace*{0.1cm}\includegraphics[width=1.0cm]{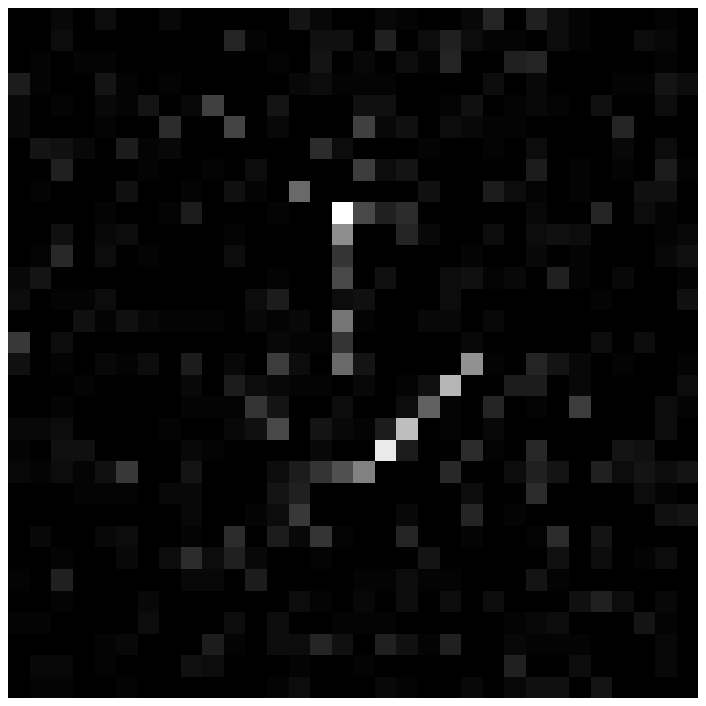}
\vspace*{-0.3cm}}
&
\parbox[c]{1cm}{\vspace*{0.1cm}\includegraphics[width=1.0cm]{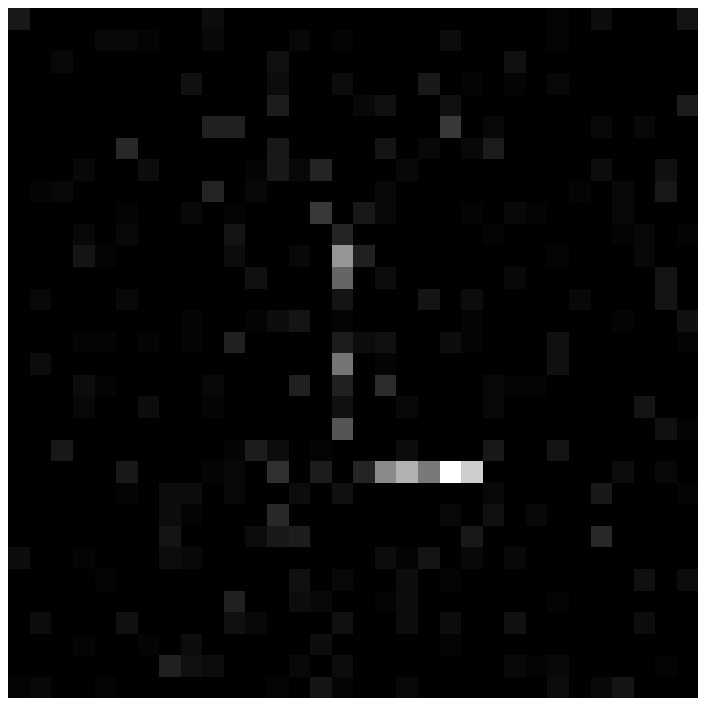}
\vspace*{-0.3cm}}
&
&
\parbox[c]{1cm}{\vspace*{0.1cm}\includegraphics[width=1.0cm]{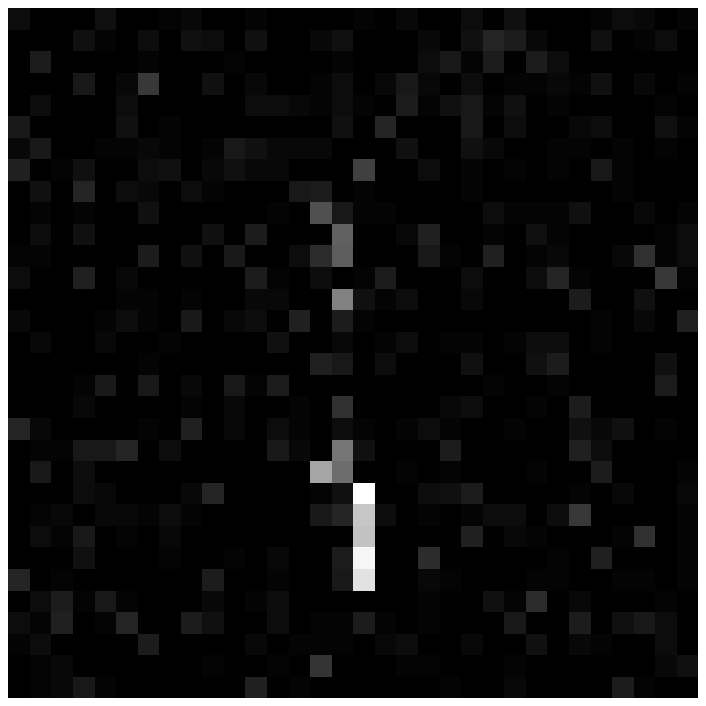}
\vspace*{-0.3cm}}  
 \\%[0.5cm]
\hline
c) &
&
&
\parbox[c]{1cm}{\vspace*{0.1cm}\includegraphics[width=1.0cm]{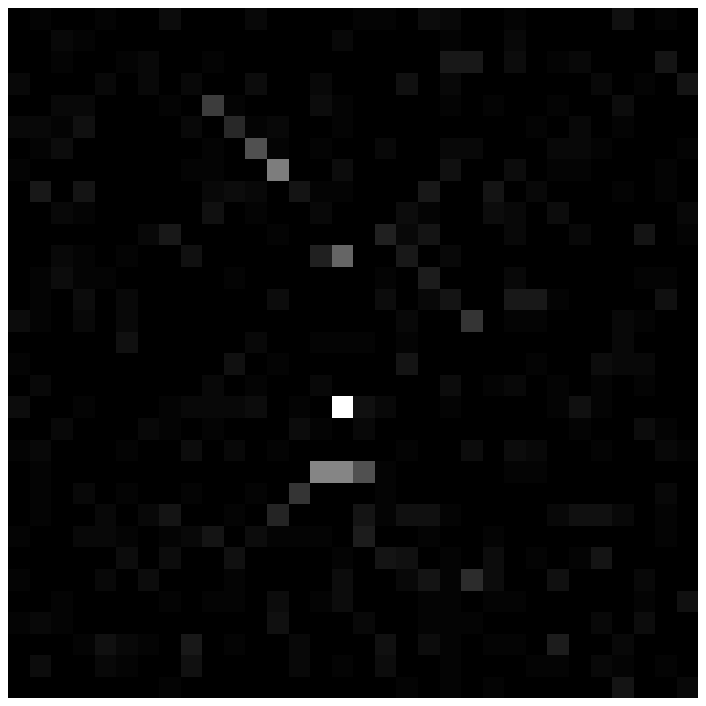}
\vspace*{-0.3cm}}
&
\parbox[c]{1cm}{\vspace*{0.1cm}\includegraphics[width=1.0cm]{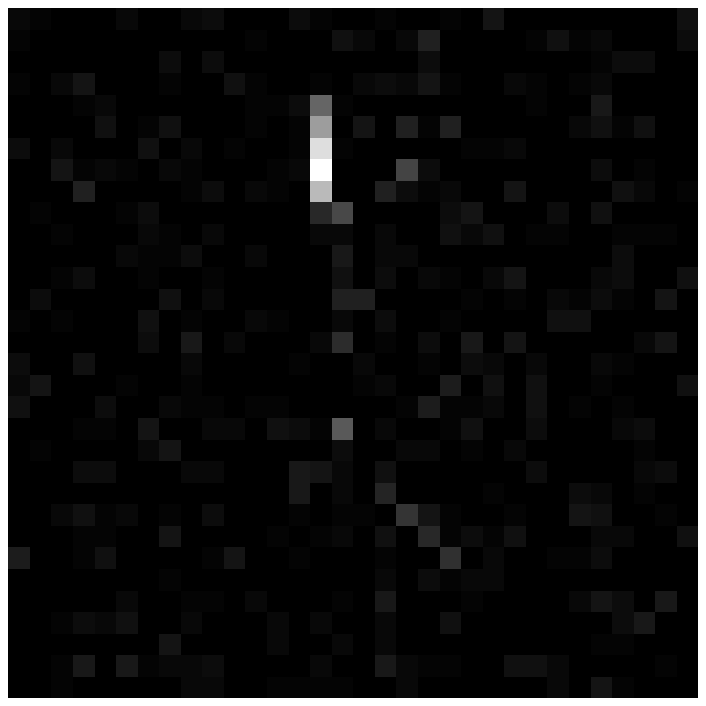}
\vspace*{-0.3cm}}
&
&
\parbox[c]{1cm}{\vspace*{0.1cm}\includegraphics[width=1.0cm]{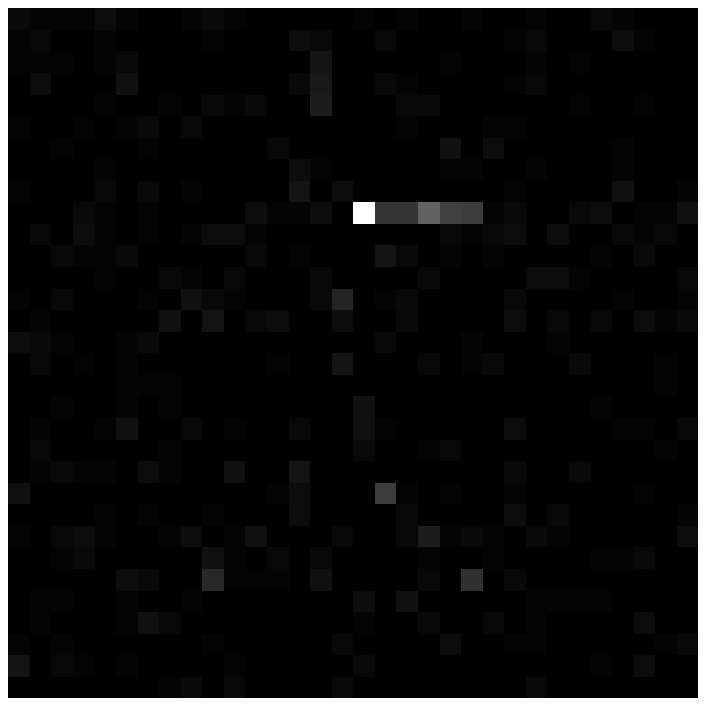}
\vspace*{-0.3cm}}
&
&
\parbox[c]{1cm}{\vspace*{0.1cm}\includegraphics[width=1.0cm]{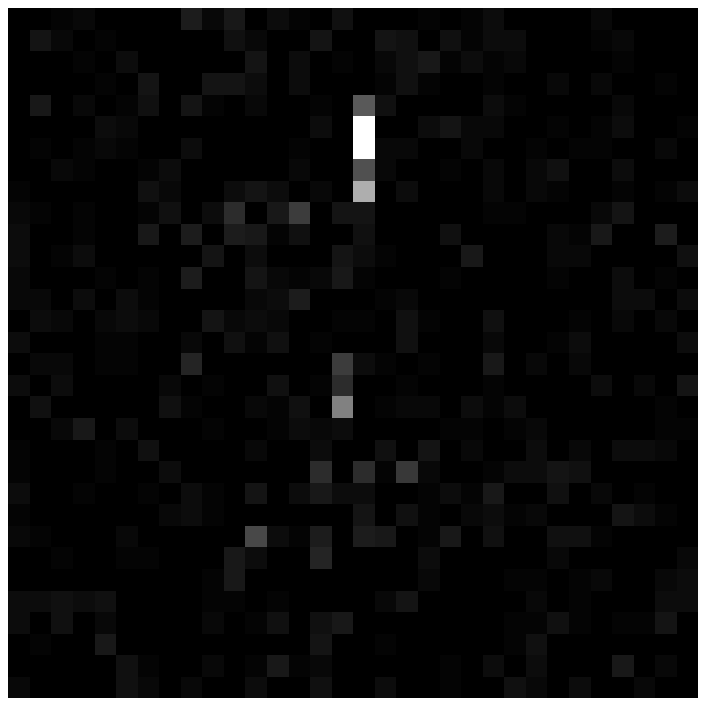}
\vspace*{-0.3cm}}
&
&
\parbox[c]{1cm}{\vspace*{0.1cm}\includegraphics[width=1.0cm]{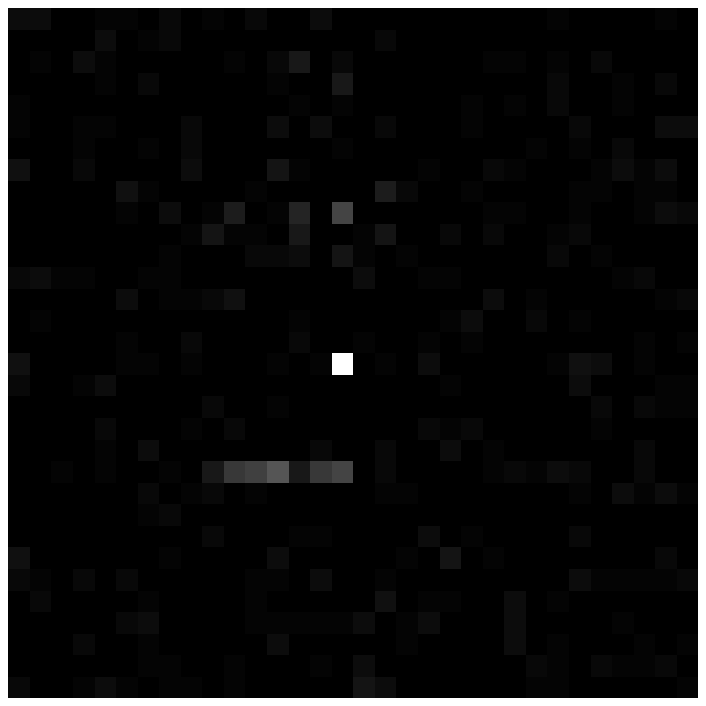}
\vspace*{-0.3cm}}
%\end{minipage}
&
\parbox[c]{1cm}{\vspace*{0.1cm}\includegraphics[width=1.0cm]{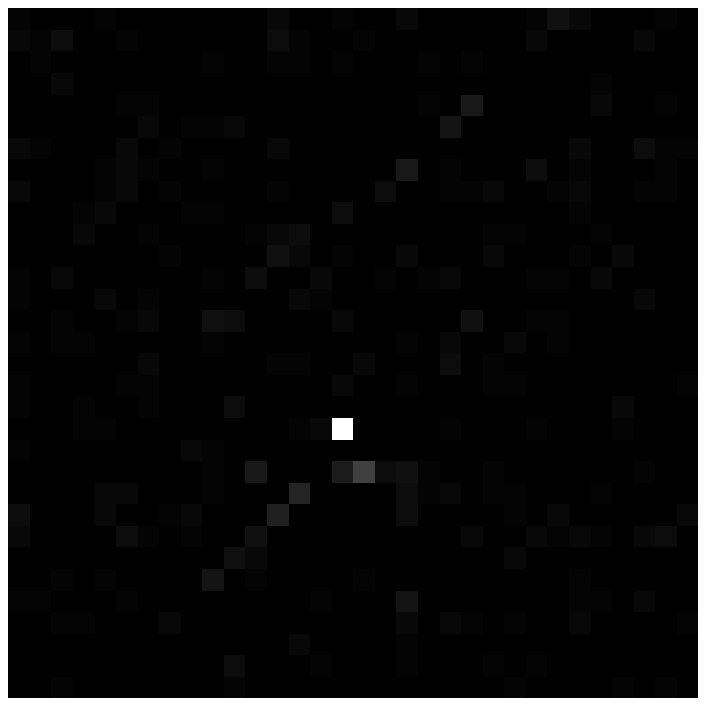}
\vspace*{-0.3cm}}
&
\parbox[c]{1cm}{\vspace*{0.1cm}\includegraphics[width=1.0cm]{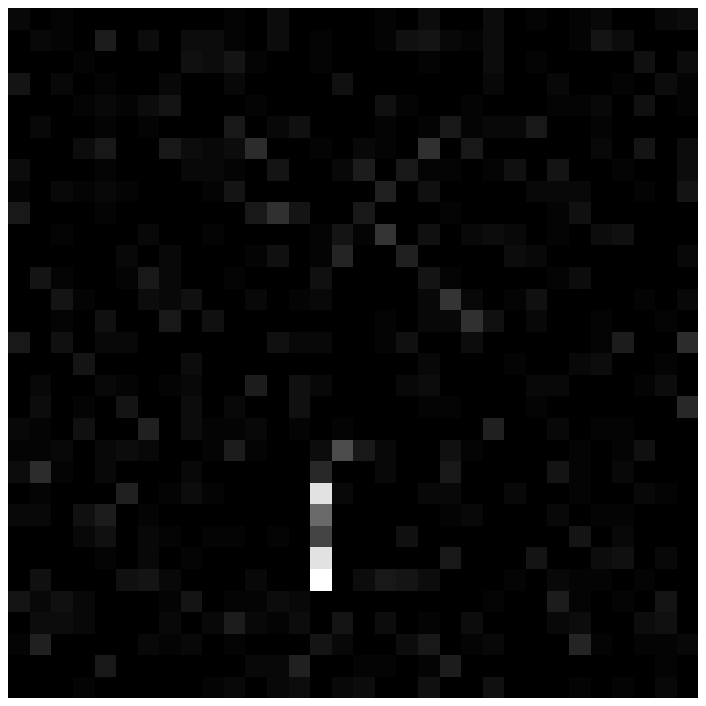}
\vspace*{-0.3cm}}
&
\parbox[c]{1cm}{\vspace*{0.1cm}\includegraphics[width=1.0cm]{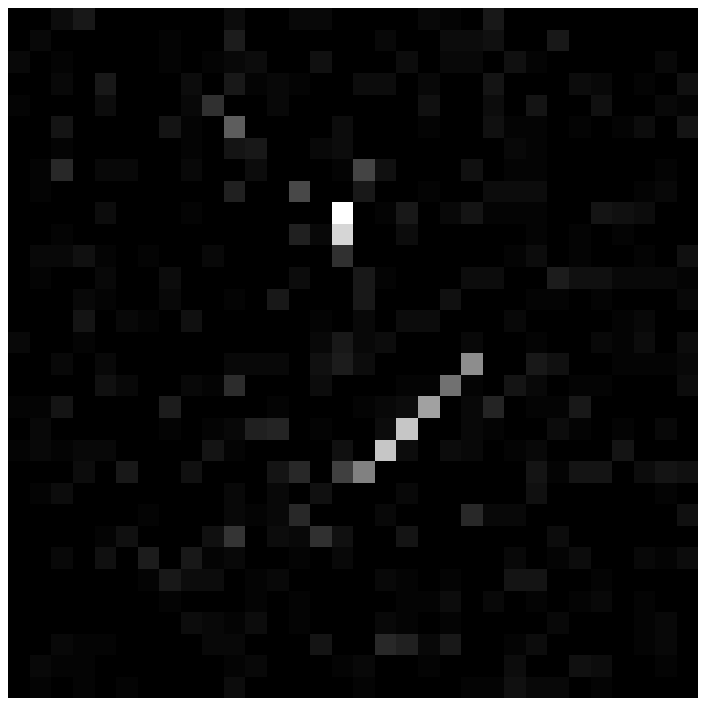}
\vspace*{-0.3cm}}
&
\parbox[c]{1cm}{\vspace*{0.1cm}\includegraphics[width=1.0cm]{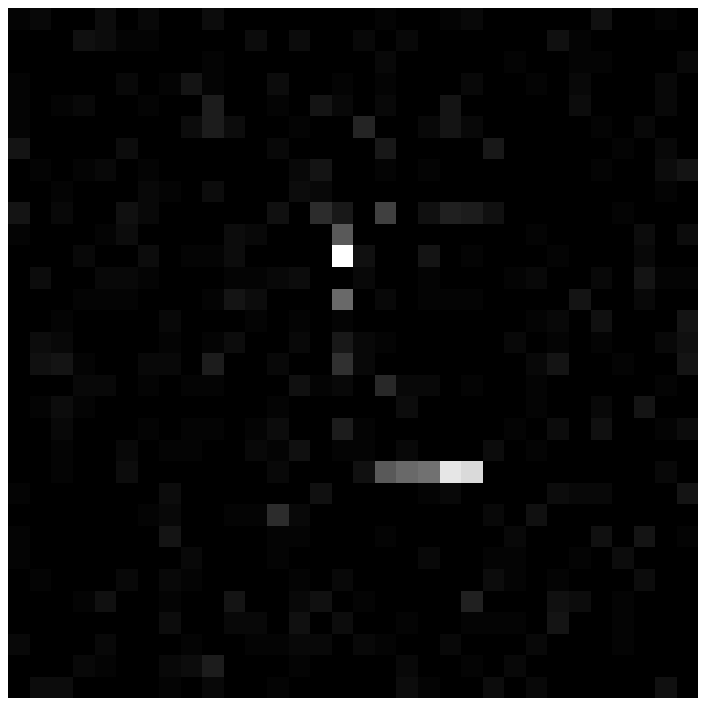}
\vspace*{-0.3cm}}
&
&
\parbox[c]{1cm}{\vspace*{0.1cm}\includegraphics[width=1.0cm]{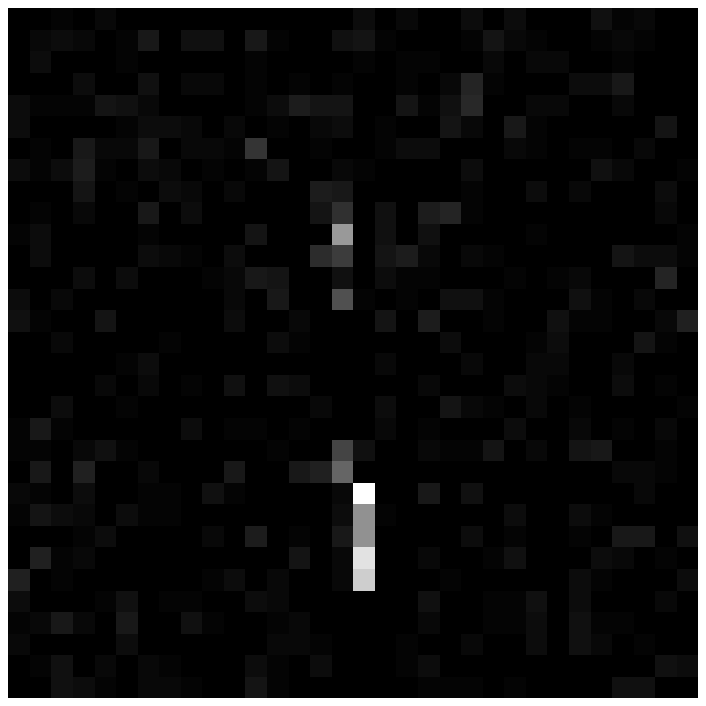}
\vspace*{-0.3cm}}
\\%[0.5cm]
\hline
\end{tabular}
\caption{Topics estimated for noisy swimmer dataset by a) proposed
  algorithm, b) LDA inference using code in \cite{Griffiths:ref}, c)
  NMF algorithm using code in \cite{betaDivergence:ref}. Topics
  closest to the 16 ideal (ground truth) topics LA1, LA2, etc.,
%  different positions of left/right arm and legs
  are shown. LDA misses $5$ and NMF misses $6$ of the ground truth
  topics while our algorithm recovers all $16$ and our topic estimates
  look less noisy.}
\label{fig:noisy_swimmer}
\end{figure*}

%%%%%%%%%%%%%%%%%%%%%%%%%%%%%%%%%

\begin{figure}[!htb]
\centering
\begin{tabular}{m{0.5cm}@{\hskip 0.05cm}m{1.1cm}@{\hskip 0.05cm}m{1.1cm}@{\hskip 0.05cm}m{1.1cm}@{\hskip 0.05cm}m{1.1cm}@{\hskip 0.05cm}m{1.1cm}@{\hskip 0.05cm}m{1.1cm}@{\hskip 0.05cm}}
%
%\hline 
%
%%%%%%%%%%%%%%%%%%%%%%
%a & a & a & a & a & a & \\
%
%%%%%%%%%%%%%%%%%%%%%
%
a) &
%
%\begin{minipage}[b]{1.0cm}
%\centerline{
\parbox[c]{1cm}{\includegraphics[width=1.0cm]{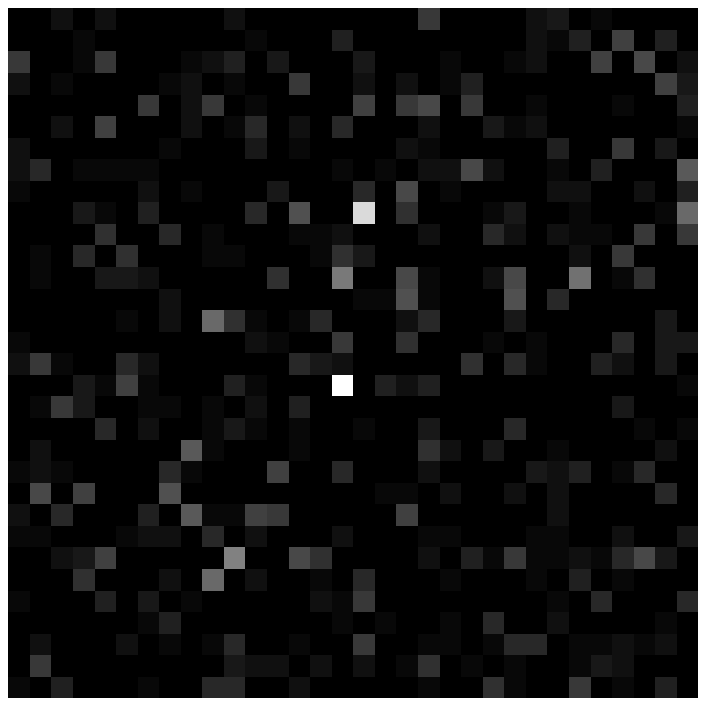}}
%\end{minipage} 
&
%
%\begin{minipage}[b]{1.0cm}
%\centerline{
\parbox[c]{1cm}{\includegraphics[width=1.0cm]{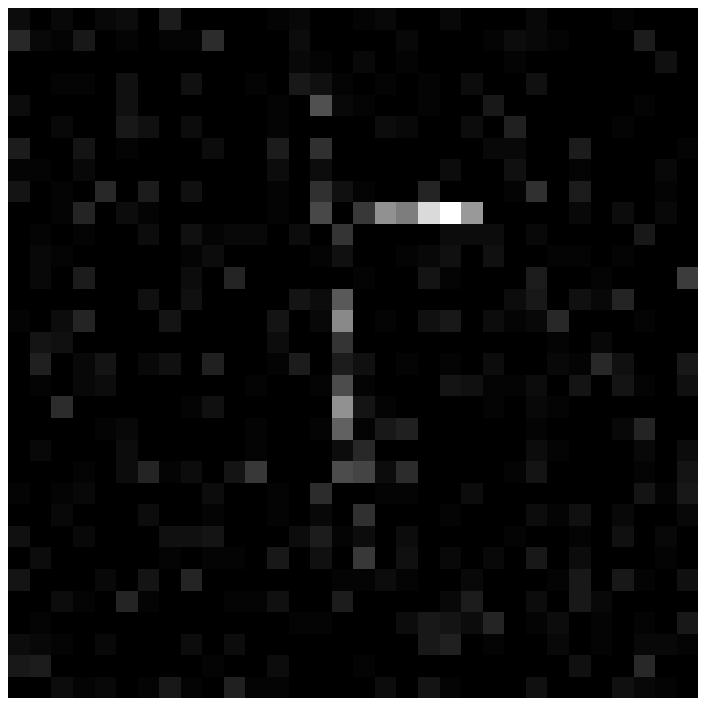}}
%\end{minipage} 
&
%
%\begin{minipage}[b]{1.0cm}
%\centerline{
\parbox[c]{1cm}{\includegraphics[width=1.0cm]{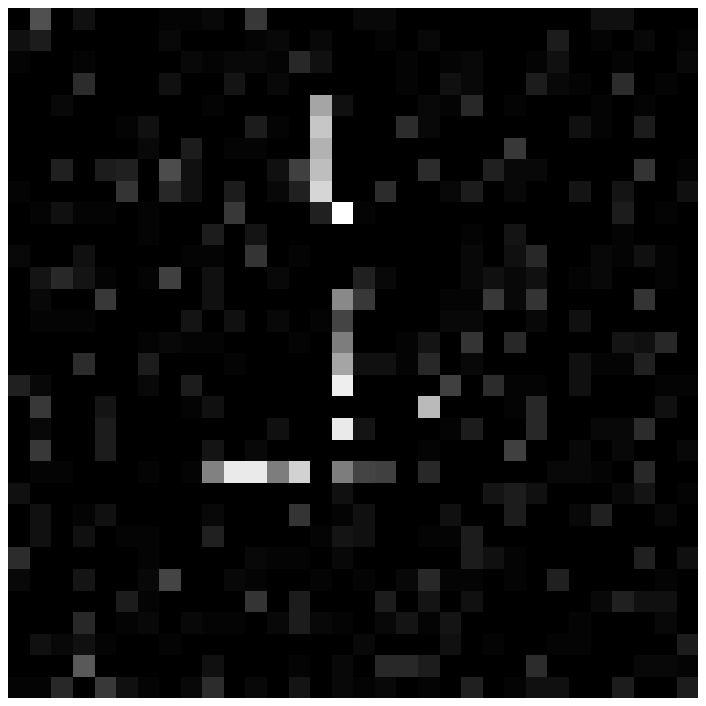}}
%\end{minipage} 
&
%
%\begin{minipage}[b]{1.0cm}
%\centerline{
\parbox[c]{1cm}{\includegraphics[width=1.0cm]{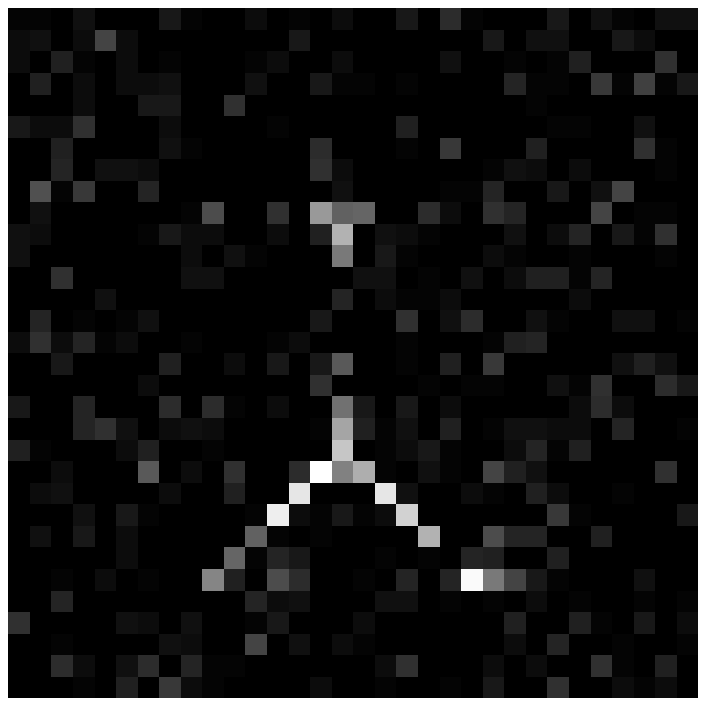}}
%\end{minipage} 
&
%
%\begin{minipage}[b]{1.0cm}
%\centerline{
\parbox[c]{1cm}{\includegraphics[width=1.0cm]{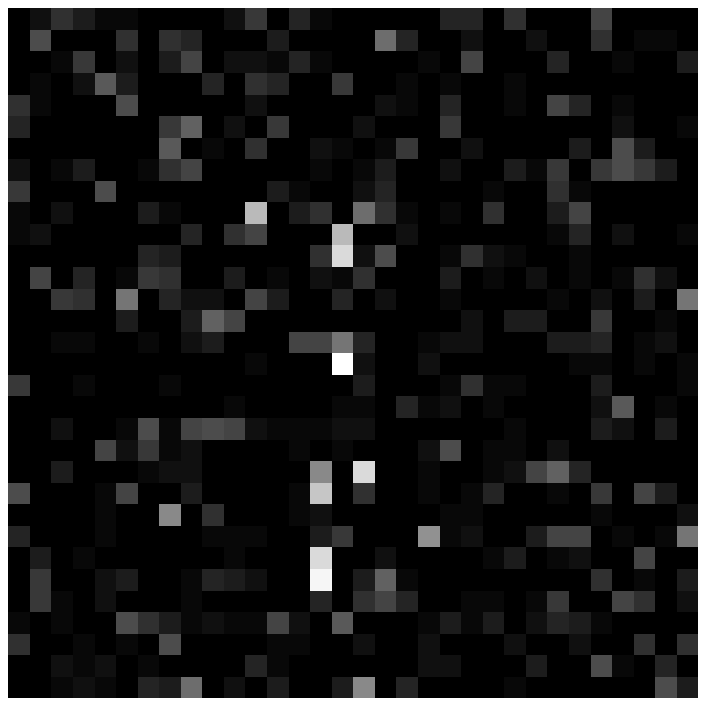}}
%\end{minipage} 
& 
\\
\\
b) &
%
%\begin{minipage}[b]{1.0cm}
%\centerline{
\parbox[c]{1cm}{\includegraphics[width=1.0cm]{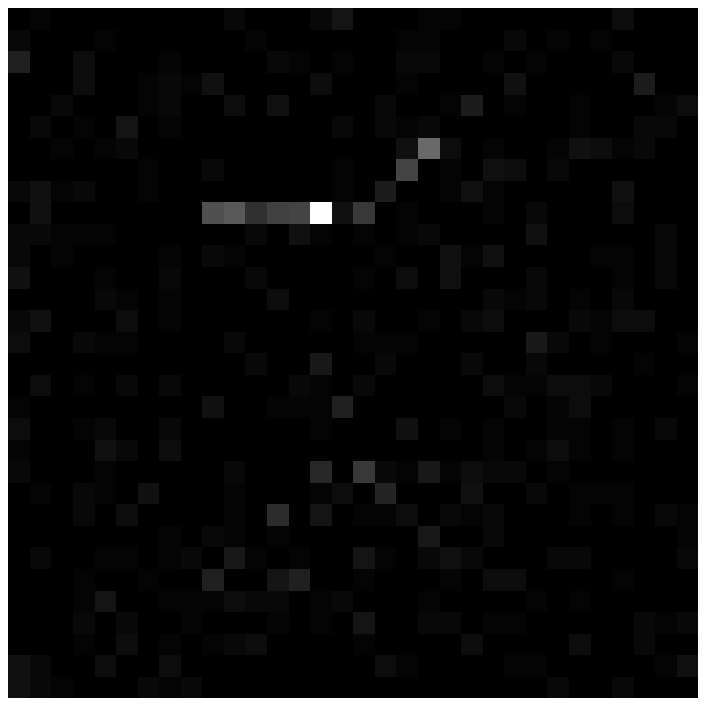}}
%\end{minipage} 
&
%
%\begin{minipage}[b]{1.0cm}
%\centerline{
\parbox[c]{1cm}{\includegraphics[width=1.0cm]{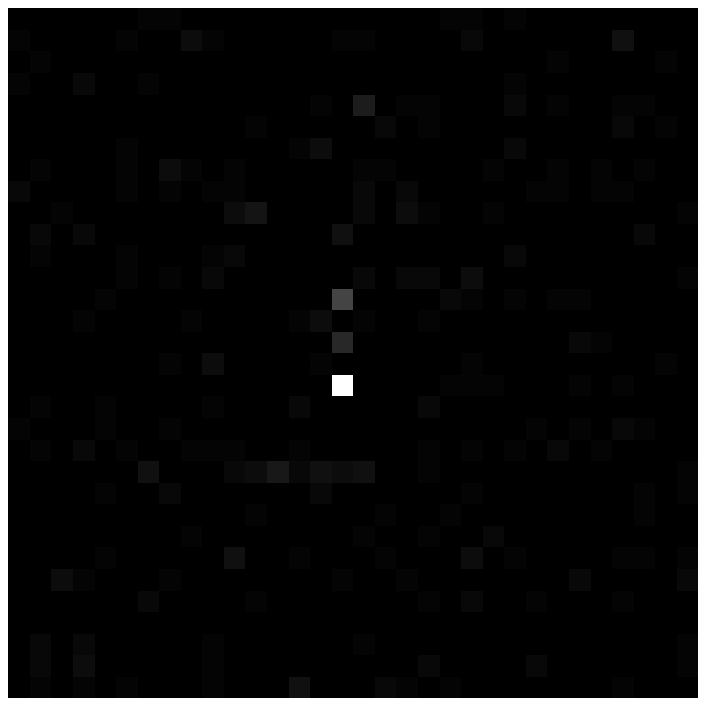}}
%\end{minipage} 
&
%
%\begin{minipage}[b]{1.0cm}
%\centerline{
\parbox[c]{1cm}{\includegraphics[width=1.0cm]{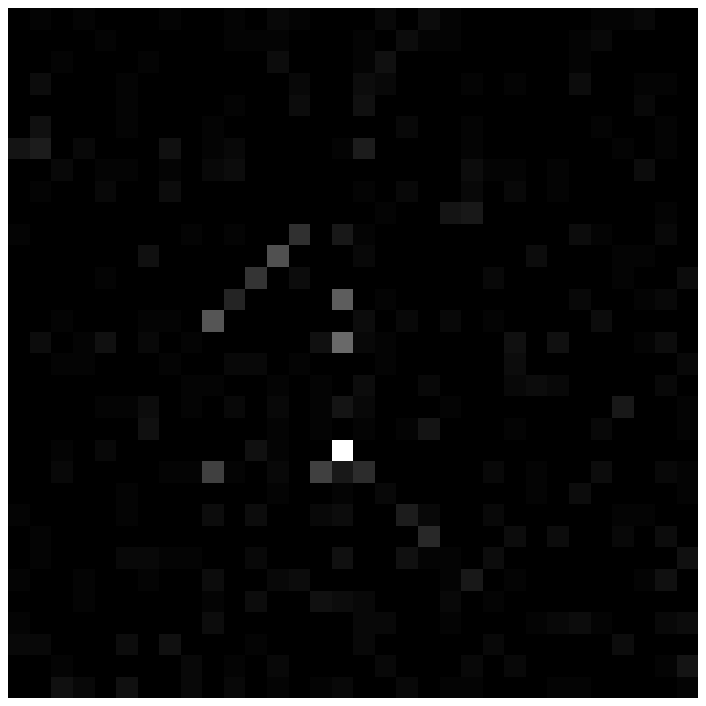}}
%\end{minipage} 
&
%
%\begin{minipage}[b]{1.0cm}
%\centerline{
\parbox[c]{1cm}{\includegraphics[width=1.0cm]{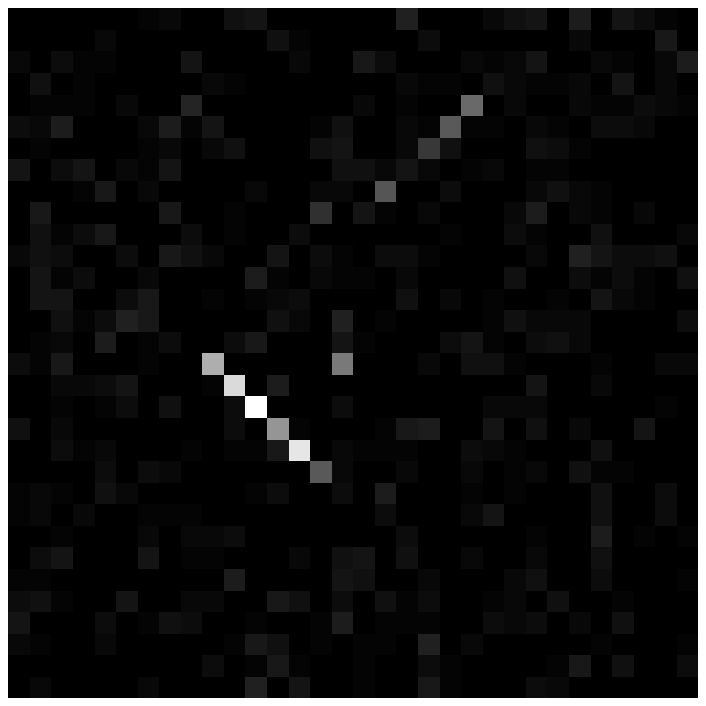}}
%\end{minipage} 
&
%
%\begin{minipage}[b]{1.0cm}
%\centerline{
\parbox[c]{1cm}{\includegraphics[width=1.0cm]{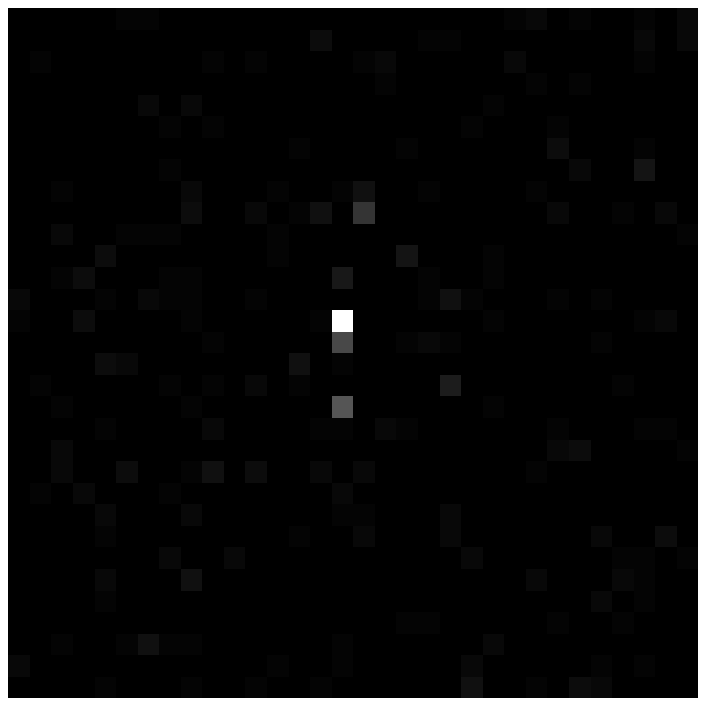}}
%\end{minipage} 
&
%
%\begin{minipage}[b]{1.0cm}
%\centerline{
\parbox[c]{1cm}{\includegraphics[width=1.0cm]{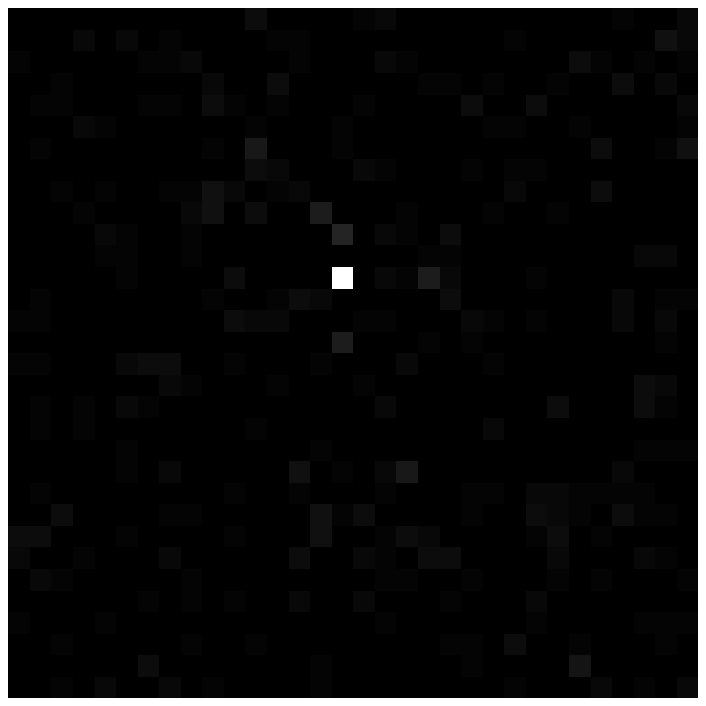}}
%\end{minipage} 
\\
%
%\hline
%
\end{tabular}
\caption{Topic errors for (a) LDA algorithm \cite{Griffiths:ref} and
  (b) NMF algorithm \cite{betaDivergence:ref} on the Swimmer
  dataset. Figure depicts topics that are extracted by LDA and NMF but
  are not close to any ``ground truth'' topic. The ground truth topics
  correspond to 16 different positions of left/right arms and legs.}
\label{fig:badimages}
\end{figure}
\vglue -1ex
%

%\begin{figure}[htb] %\vfill
%\begin{minipage}[b]{1.0\linewidth}
%  \centering
% \centerline{\includegraphics[width=10.0cm]{swimmer_result.eps}}
%\end{minipage}
%
%\end{figure}

% intro to the dataset
In this section we apply our algorithm to the synthetic
\textit{swimmer} image dataset introduced in
\cite{Donhunique:ref}. There are $M = 256$ binary images each of
$W=32\times32=1024$ pixels. Each image represents a swimmer composed
of four limbs, each of which can be in one of $4$ distinct positions,
and a torso.

% Interpret as topic modeling
We interpret pixel positions $(i,j), 1\leq i,j\leq 32$ as words in a
dictionary. Documents are images, where an image is interpreted as a
collection of pixel positions with non-zero values. Since each of the
four limbs can independently take one of four positions, it turns out
that the topic matrix $\beta$ satisfies the separability assumption
with $K = 16$ ``ground truth'' topics that correspond to $16$ {\it
  single} limb positions.
% experiment setting
Following the setting of \cite{betaDivergence:ref}, we set body pixel
values to 10 and background pixel values to 1. We then take each
``clean'' image, suitably normalized, as an underlying distribution
across pixels and generate a ``noisy'' document of $N = 200$ iid
``words'' according to the topic model. Examples are shown in
Fig.~\ref{fig:sample_swimmer}. We then apply our algorithm to the
``noisy'' dataset. We again compare our algorithm against LDA and the
NMF algorithm from \cite{betaDivergence:ref}. Results are shown in
Figures \ref{fig:noisy_swimmer} and \ref{fig:badimages}. Values of
tuning parameters $\lambda_1, \gamma$, and $\lambda_2$ are set as in
Sec.~\ref{sec:synthetic}. Specifically, $\lambda_1 = 0.1, \lambda_2 =
0.01$ for the results in Figs.~\ref{fig:noisy_swimmer} and
\ref{fig:badimages}.

This dataset is a good validation test for different algorithms since
the ground truth topics are known and are unique. As we see in
Fig.~\ref{fig:badimages}, both LDA and NMF produce topics that do not
correspond to any {\it pure} left/right arm/leg positions. Indeed,
many estimated topics are composed of multiple limbs. Nevertheless, no
such errors are realized in our algorithm and our topic-estimates are
closer to the ground truth images.

%\ref{fig:badimages}, some estimated topics are composed of more than
%one limbs even if we set sufficient number of iterations.

\vspace{-4pt}

\subsection{Text Corpora}

% table for NIPS data
%
\begin{table}[!htb]
\centering
\begin{tabular}{|c|c|c|c|}
%{|p{0.18\linewidth}|p{0.16\linewidth}| p{0.16\linewidth}| p{0.16\linewidth}|}
%
\hline
``\bf{chips}'' & ``\bf{vision}'' & ``\bf{networks}'' & ``\bf{learning}'' \\
\hline
chip & visual & network & learning \\
circuit & cells &  routing & training \\
analog & ocular & system & error \\
current & cortical & delay & SVM \\
gate & activity & load & model \\
\hline
\end{tabular}
%
%\label{NIPS:table}
%
\centerline{}\medskip \\
\centering
\begin{tabular}{|c|c|c|c|}
%{|p{0.16\linewidth}|p{0.16\linewidth}| p{0.16\linewidth}| p{0.16\linewidth}|}
%
\hline
``\bf{election}''  & ``\bf{law}'' & ``\bf{market}'' &``\bf{game}'' \\
\hline
 state  & case & market & game \\
 politics & law & executive & play \\
 election & lawyer & industry & team \\
 campaign & charge & sell & run  \\
 vote & court &	business & season \\
%
% party & cup & rule & share & win \\
%
% Democrat & oil & investigation & bank & player \\
%
% support & dinner & judger & trade & score \\
%
\hline
\end{tabular}
\caption{Most frequent words in examples of estimated topics. Upper:
  \textit{NIPS}, with $K=40$ topics; Lower: \textit{NY Times},
  with $K=20$ topics}
\label{topicword:table}
\end{table}
In this section, we apply our algorithm on two different text corpora,
namely, the NIPS dataset \cite{NIPSdataset:ref} and the \textit{New
  York (NY) Times} dataset \cite{NYT3K:ref}.
%
%http://ai.stanford.edu/~gal/data.html is the site of NIPS dataset.
% http://www.princeton.edu/~achaney/tmve/nyt_demo/browse/topic-presence.html for the NYT news datasets.
%
In the NIPS dataset, there are $M=2484$ documents with $W=14036$ words
in the vocabulary. There are, on average, $N \approx 900$ words in
each document. In the NY Times dataset, $M=3000$, $W=9340$, and $N
\approx 270$. The vocabulary is obtained by deleting a standard
``stop'' word list used in computational linguistics, including
numbers, individual characters, and some common English words such as
``the''. Words that occur less than $5$ times in the dataset and the
words that occur in less than $5$ documents are removed from the
vocabulary as well. The tuning parameters $\lambda_1, \gamma,$ and
$\lambda_2$ are set in the same way as in Sec.~\ref{sec:synthetic}
(specifically, $\lambda_1 = 0.1$ and $\lambda_2=0.1$).

Table~\ref{topicword:table} depicts typical topics extracted by our
algorithm. For each topic we show its most frequent words, listed in
descending order of estimated probability. Although there is no
``ground truth'' to compare with, the most frequent words in the
estimated topics do form recognizable themes. For example, in the NIPS
dataset, the set of (most frequent) words ``chip'', ``circuit'', etc.,
can be annotated as ``IC Design''; The words ``visual'', ``cells'',
etc., can be labeled as ``human visual system''. As a point of
comparison, we also experimented with related convex programming
algorithms \cite{ARORA:ref,recht2012factoring} that have recently
appeared in the literature. We found that they fail to produce
meaningful results for these datasets.

%\section{Conclusion}
%%
%In this paper we studied the probabilistic topic modeling problem in
%the view of nonnegative matrix factorization. We demonstrated that
%under the separability and sparsity assumption, the problem has nice
%geometric structures robust to noise. We proposed an approach that
%explore the properties. Our algorithm could avoid local search, runs
%in polynomial-time and can deal with dataset without huge number of
%samples in practice. It is validated by different synthetic and real
%world dataset.

%\vfill
%\pagebreak

\bibliographystyle{IEEEbib}
%\bibliography{strings,refs}
\bibliography{refs}

\end{document}